\documentclass[10pt,twocolumn,letterpaper]{article}

\usepackage{cvpr}
\usepackage{times}
\usepackage{epsfig}
\usepackage{graphicx}
\usepackage{amsmath,calc}
\usepackage{amssymb,mathtools}
\usepackage{amsfonts}
\usepackage{color}
\usepackage{verbatim}
\usepackage{lipsum}
\usepackage{multirow,bigdelim}
\usepackage{rotating}
\usepackage{hhline}
\usepackage{mwe}
\usepackage{graphbox}
\usepackage{transparent}
\usepackage[ruled,titlenotnumbered,linesnumbered,noline]{algorithm2e}
\usepackage{algorithmic}
\usepackage[hang,flushmargin]{footmisc}

\def\labelvars{\mathbf{f}}
\def\labelvar{f}
\def\clabelvars{\mathbf{f}'}
\def\clabelvar{f'}

\def\pixelset{\Omega}
\def\neighbset{\mathcal{N}}

\def\labelset{\mathcal{L}}
\def\TConst{T}
\def\Tree{\mathcal{T}}
\def\margins{\delta}

\def\unary{D}
\def\parent{\mathcal{P}}
\def\pairwise{V}

\def\a{a}
\def\labelstyle{\small }
\def\gpath[#1,#2]{\Gamma({#1},{#2})}

\usepackage[pagebackref=true,breaklinks=true,letterpaper=true,colorlinks,bookmarks=false]{hyperref}

\cvprfinalcopy 

\def\cvprPaperID{539} 

\ifcvprfinal\fi
\begin{document}



\title{\vspace{-4ex}Efficient optimization for Hierarchically-structured Interacting Segments (HINTS)\\ \vspace{-2ex}}

\author{
Hossam Isack$^1$ \hspace{3ex}Olga Veksler$^1$ \hspace{3ex} Ipek Oguz$^2$ \hspace{3ex} Milan Sonka$^3$  \hspace{3ex} Yuri Boykov$^1$ \vspace{1ex} \\
$^1$Computer Science, University of Western Ontario, Canada\\
$^2$Department of Radiology, University of Pennsylvania, USA \\
$^3$Electrical and Computer Engineering, University of Iowa, USA\vspace{-1ex}}

\maketitle
\begin{abstract}
\vspace{-1ex}
We propose an effective optimization algorithm for a general
hierarchical segmentation model with geometric
interactions between segments.
Any given tree can specify a partial
order over object labels defining a hierarchy.
It is well-established that segment interactions, such as inclusion/exclusion and margin constraints,
make the model significantly more discriminant.
However, existing optimization methods do not allow full use of such models. \linebreak
Generic a-expansion  results in weak local minima, while
common binary multi-layered formulations lead to non-submodularity, complex high-order potentials, or polar domain unwrapping and shape biases.
In practice, applying these methods to arbitrary trees does not work except for simple cases.
Our main contribution is an optimization method for the Hierarchically-structured Interacting Segments (HINTS) model with arbitrary trees. Our Path-Moves algorithm
is based on multi-label MRF formulation and can be seen as a combination of well-known a-expansion and Ishikawa techniques. We show state-of-the-art biomedical segmentation for many diverse examples of complex trees.
\vspace{-3ex}
\end{abstract}

\section{Introduction}
\label{sc:intro}
Basic cues like smooth boundaries and appearance models are often insufficient to regularize complex segmentation problems. This is particularly true in medical applications where objects have weak contrast boundaries and overlapping appearances.
Thus, additional priors are needed, e.g.~shape-priors \cite{olga:ECCV08}, volumetric constraints \cite{volumetricbias15}, or segments interaction \cite{xiaodong:logismos,DB:ICCV09}.
The latter constraint is the essence of {\bf H}ierarchically-structured {\bf Int}eracting {\bf S}egments (HINTS) model\footnote{HINTS model was proposed by \cite{DB:ICCV09} by was not given a name.} \cite{DB:ICCV09,xiaodong:logismos}, which was successfully applied to many segmentation problems, e.g.~cells 
\cite{pfuamitochondria}, joint cartilage \cite{xiaodong:logismos}, cortical \cite{oguz2014logismos} or tubular \cite{li2006optimal} surfaces.

\begin{figure}[h]
\begin{center}
\begin{tabular}{@{\hspace{0ex}}cc}
\includegraphics[width=0.27\linewidth]{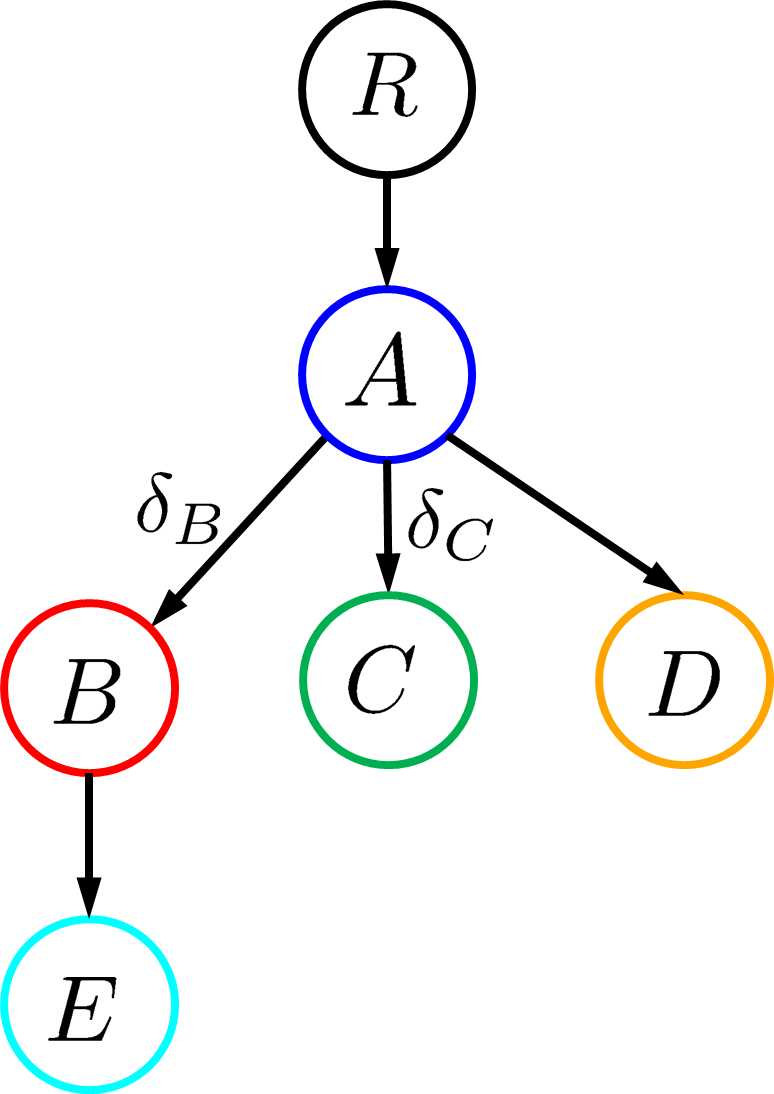}&
\includegraphics[width=0.55\linewidth]{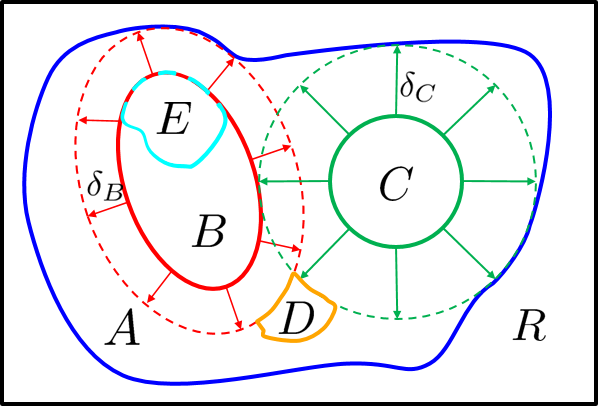}\\
(a) tree $\Tree$ \& margins $\margins$ & (b) a feasible segmentation
\end{tabular}
\end{center}
\caption{\labelstyle (a) tree with 6 labels, and $\margins_B$ and $\margins_C$ are min-margins of labels B and C, respectively. None displayed margins imply zero min-margin. (b) feasible segmentation that satisfies the hierarchical-structure, i.e.~partial ordering, and margins defined by $\Tree$ and $\margins$. Notice how B's outside boundary pushes its parent's and siblings' outside boundaries to be at least $\margins_B$ pixels away from it.}
\label{fig:basictree}
\end{figure}

\vspace{-3ex}
\paragraph{HINTS model overview:\!\!\!} Any hierarchically-structured\footnote{We use hierarchically-structured and partially ordered interchangeably.} segments could be represented as a label tree $\Tree$,
see Fig.~\ref{fig:basictree}. Tree $\Tree$ defines topological relationship between segments as follows; (a) child-parent
relation means the child segment is inside its parent's segment, (b) sibling relation means the corresponding segments exclude each other.
For example, in Fig.~\ref{fig:basictree} $A$ is inside $R$, $E$ is inside $B$, while $B$, $C$ and $D$ exclude each other in $A$.
Min-margin is one form of interaction between regions. If  region $X$ has $\margins_X$ min-margin then its outside boundary pushes away the outside boundary of its parent and siblings by at least $\margins_X$, Fig.\ref{fig:basictree}(b).

\begin{figure}[h]
\begin{center}
\begin{tabular}{@{\hspace{0ex}}c@{\hspace{0.5ex}}c@{\hspace{0.5ex}}c@{\hspace{0.5ex}}c}
\includegraphics[width=0.24\linewidth]{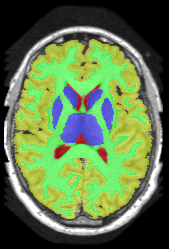}&
\includegraphics[width=0.24\linewidth]{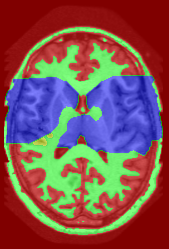}&
\includegraphics[width=0.24\linewidth]{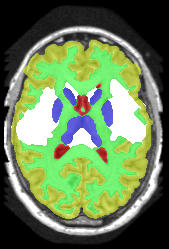}&
\includegraphics[width=0.24\linewidth]{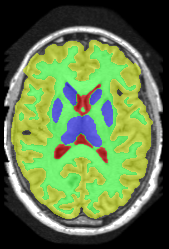}\\
{\small ground truth}&{\small  \a-exp\cite{BVZ:PAMI01,DB:ICCV09}}&
{\small QPBO \cite{rotherQPBO,DB:ICCV09}} & {\small ours}
\end{tabular}
\end{center}
\caption{{\labelstyle HINTS segmentation of brain using different optimization methods where white-matter (yellow), grey-matter (green) and background are nested regions, and cerebrospinal fluid (red) and sub-cortical grey-matter (blue) are mutually exclusive regions inside white-matter. Starting from a trivial solution \a-exp converged to a bad local minimum unlike Path-Moves which explores more solutions. QPBO failed to label some pixels (shown in white) due to the non-submodular energy and ambiguous color models.}}
\label{fig:teaser}
\end{figure}

\paragraph{Limitations of previous algorithms:\!\!\!} We extend \cite{DB:ICCV09}, which introduced HINTS for arbitrary trees.
In \cite{DB:ICCV09} \a-expansion (\a-exp) \cite{BVZ:PAMI01} was used to optimize the multi-label formulation of HINTS, but it often results in bad local minima due to complexities of interaction constraints, e.g.~Fig.\ref{fig:teaser}.
The contribution of \cite{DB:ICCV09} is a binary multi-layered HINTS formulation.
They use high-order data terms, which are not easy to convert into unary and pairwise potentials for arbitrary trees.
Their algorithm's global optimality guarantee depends on the tree at hand.
Only trees that do not yield {\em frustrated cycles} \cite{rotherQPBO} have this guarantee, but this is not immediately obvious for any given tree.
In \cite{DB:ICCV09}, non-submodular binary energy implied by frustrated cycles were addressed by QPBO \cite{rotherQPBO}.
In practice, QPBO produces only partial solutions for most trees, see Figs.~\ref{fig:teaser}, \ref{fig:brainresults} and \ref{fig:heartresult}.

As an alternative to QPBO, \cite{ulen2013efficient} formulated HINTS as constraint optimization. They solve the Lagrangian dual of this NP-hard problem using an iterative sequence of graph cuts. However, the duality gap may be arbitrarily large and the optimum for HINTS is not guaranteed. Their super-gradient optimization of Lagrange multiplier guesses initial solutions and time-step parameters.
Also, according to Lemma 5 in \cite{ulen2013efficient} their super-gradient corresponds to the hard exclusion constraint of HINTS, which is $\{0,\infty\}$-valued.
They do not discuss how this affects the algorithm.

In \cite{xiaodong:logismos} the authors generalize  their earlier method \cite{li2006optimal} for segmenting multiple nested surfaces, i.e.~$\Tree$ is a chain.
In \cite{xiaodong:logismos} the aim was to segment multiple mutually exclusive objects each with a set of nested surfaces, i.e.~$\Tree$ is a spider\footnote{Tree with one node of degree $\ge 3$ and all others with degree $\le 2$.}.
But, the proposed approach can handle only a single pair of mutually exclusive objects in a given image region. As such \cite{xiaodong:logismos} requires a prior knowledge of the region of interaction for two excluded objects,
or computes it using a problem specific trained classifier \cite{xiaodong:logismos}. Such prior knowledge is not required for our method.
In contrast to our approach, \cite{xiaodong:logismos} requires a sufficiently close initial segmentation satisfying interaction constraints. In all of our experiments we started from a trivial solution. Unlike our approach, \cite{xiaodong:logismos} implicitly imposes a star like shape prior \cite{olga:ECCV08} and use non-homogeneous anisotropic polar grids.

If interactivity (min-margin) constraints are dropped HINTS degenerates to {\em tree-metric labeling}.
Certain tree-metric labeling problems are addressed in \cite{PedroGobalTreeMetric} using DP to find the global optima if the data terms are also a tree-metric.
Recently, \cite{Baxter2017} used convex relaxation to approximate labeling problems where labels are leafs of
a DAG. This problem can be reduced\footnote{Personal communication with the authors of \cite{Baxter2017}.} to
general {\em metric labeling} \cite{eva2002}. Such labeling problems are significantly different from HINTS due to interactions between segments.

\vspace{-2.5ex}
\paragraph{Motivation for Path-Moves:\!\!\!}
In the context of multi-label HINTS formulation, we propose an effective move-making algorithm
applicable to {\em arbitrary} label trees  $\Tree$  avoiding limitations of the previous optimization methods.

In contrast to \a-exp \cite{BVZ:PAMI01}, our Path-Moves are non-binary: when expanding label $\alpha$ any pixel can change its current label to any label along the path connecting its current label and $\alpha$ in the tree.
Optimization uses our generalization of the well-known multi-layered Ishikawa technique \cite{Ishikawa:PAMI03}
for convex potentials over strictly ordered labels.
In essence, Path-Moves combine
\a-exp and Ishikawa. In the special case of a chain-tree our algorithm reduces to Ishikawa-like construction in \cite{DB:ICCV09,Ishikawa:PAMI03}
finding global minimum in one step. On the contrary, when $\Tree$ is a single-level star
our algorithm reduces to \a-exp. Note that closely related multi-label {\em range-moves}
\cite{olgarangemoves:IJCV12} also combine \a-exp and
Ishikawa for non-convex pairwise potentials over strictly ordered labels (a chain).
In contrast to Path-Moves, in {\em range-moves} all pixels have the same set of feasible labels.

Our contributions are summarized below:
\vspace{-1.5ex}
\begin{itemize}
\item we propose Path-Moves - approximate optimization method applicable to HINTS.
Unlike \cite{DB:ICCV09,xiaodong:logismos}, Path-Moves work for  arbitrary  trees avoiding weak local minima typical of \a-exp~\cite{BVZ:PAMI01}  in the context of HINTS.
\vspace{-1ex}
\item we show how a generalization of star shape priors, e.g.~\cite{olga:ECCV08,gulshan2010geodesic,isackhedgehog}, integrate into multi-label HINTS model, if needed. Path-Moves can address this too.
\vspace{-3.5ex}
\item we show state-of-the-art biomedical segmentation results for complex trees.
\vspace{-1ex}
\end{itemize}

\section{\fontsize{11.195}{11.195}\selectfont{Hierarchically-structured Interacting Segments}}
\label{sc:segwithtopologicalconst}
Given pixel set $\pixelset$, neighborhood system $\neighbset$, and labels (regions) $\labelset$ the HINTS model can be formulated as
\begin{equation} \label{eqn:OurMainEnergy}
\!\!\!\!\!\!E(\labelvars)= \overbrace{\sum_{p \in \pixelset} \!\!\unary_p(\labelvar_p)}^\text{\small {\em data}} + \overbrace{\lambda\!\!\sum_{pq \in \neighbset}\!\!\pairwise_{pq}(\labelvar_p,\labelvar_q)}^\text{\small {\em smoothness}} + \overbrace{\phantom{\sum}\!\!\!\!\!\!\!\TConst(\labelvars)}^{\substack {{\text{\small {\em interaction}}} \\ {\text {\small {\em constraints}}}}}
\end{equation}
where $\labelvar_p$ is a label assigned to $p$ and \mbox{$\labelvars=\left[\labelvar_p \in \labelset |\; \forall p \in \pixelset \right]$} is a labeling of all pixels.

The {\em data}  and {\em smoothness} terms are widely used in segmentation, e.g.~\cite{BVZ:ICCV99,BJ:01}. Data term $\unary_p(\labelvar_p$) is the cost incurred  when pixel $p$ is assigned to label $\labelvar_p$. Usually, $\unary_p$ is negative log likelihood of the label's probabilistic model, which could be fitted using scribbles \cite{BJ:01,GrabCuts:SIGGRAPH04} or known a priori.

The {\em smoothness} term  regularizes segmentation discontinuities. A discontinuity occurs when two neighboring pixels $(p,q)\in \neighbset$ are assigned to different labels. Parameter $\lambda$ weights the importance of the smoothness term.
The most commonly used {\em smoothness} potential is Potts model \cite{BVZ:PAMI01}.\break
We use tree-metric smoothness \cite{PedroGobalTreeMetric,gridchyn2013potts} which is more true to the physical structure of the labels in some settings, especially medical segmentation as we explain shortly.

\newpage A function $V$ is tree-metric if there exists a tree with non-negative edge weights and $V(u,w)$ is equal to the sum of edge weights along the unique path between nodes $u$ and $w$ in the tree. In our setting the label tree $\Tree$ is such a tree and $\pairwise_{pq}$ is completely defined by assigning non-negative weights to every edge in $\Tree$. Thus, for any $\alpha$, $\beta$ in $\labelset$
\begin{equation}
\pairwise_{pq}(\alpha,\beta) =
\underset{ij \in\gpath[\alpha,\beta]}{\sum}
\pairwise_{pq}(i,j),
\label{eqn:treemetriccond}
\end{equation}  where $\gpath[\alpha,\beta]$ is the set of ordered labels on the path between $\alpha$ and $\beta$ in the undirected tree $\Tree$. The summation in \eqref{eqn:treemetriccond} is between pairs of neighboring labels on path $\gpath[\alpha,\beta].$

To motivate tree-metric smoothness consider $\Tree$ in Fig.\ref{fig:basictree}(a).
This tree implies that regions $R$, $A$ and $D$ are nested. For example, $R$, $A$ and $D$ could be background, cell and nucleus, respectively.
In the physical world boundaries of nested regions never merge into a single boundary.
That is, if in the image we observe a boundary between $D$ and $R,$ this corresponds to two boundaries, namely $D/A$ and $A/R$ in the physical world. Therefore, the $D/R$ boundary cost should be the sum of $D/A$ and $A/R$ boundary costs.
The summation property of nested boundaries can be modeled as tree-metric smoothness.
In contrast, Potts model penalizes multiple nested boundaries as a single boundary.

\begin{figure}[t]
\begin{center}
\begin{tabular}{cc}
\includegraphics[align=c,width=0.3\linewidth]{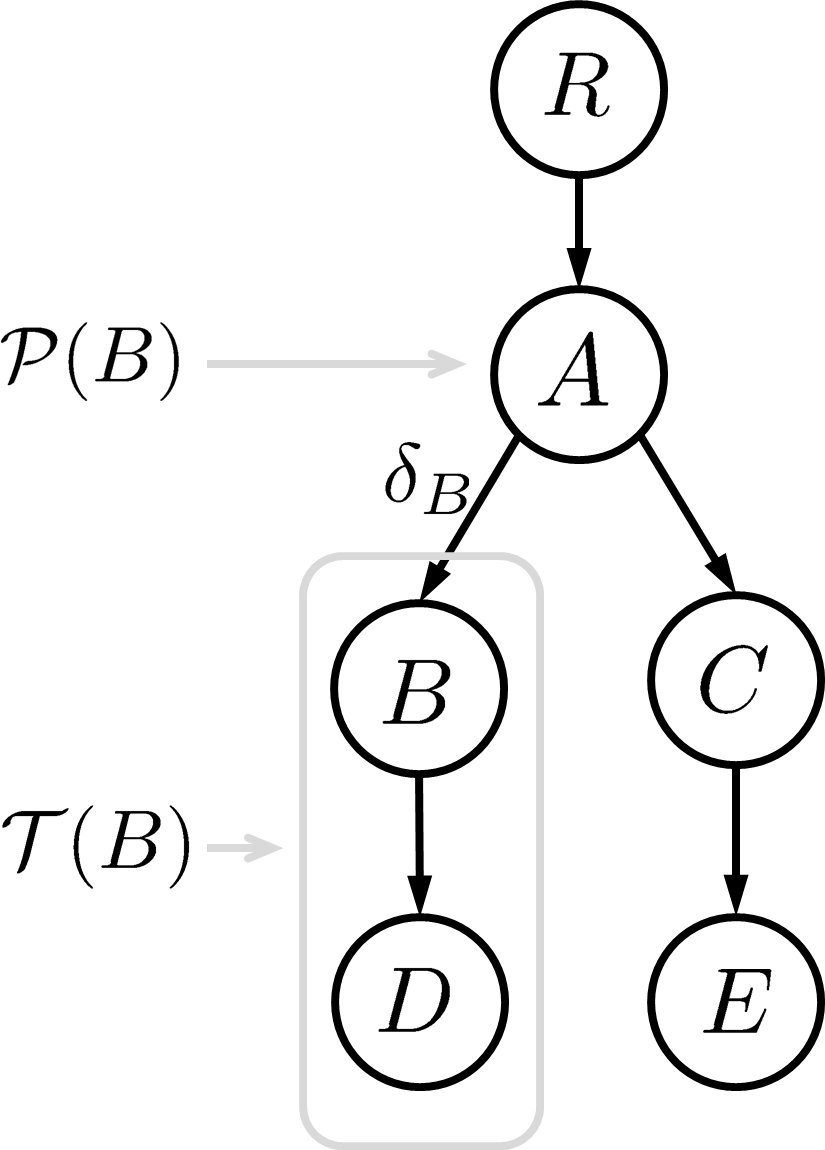}&
\includegraphics[align=c,width=0.5\linewidth]{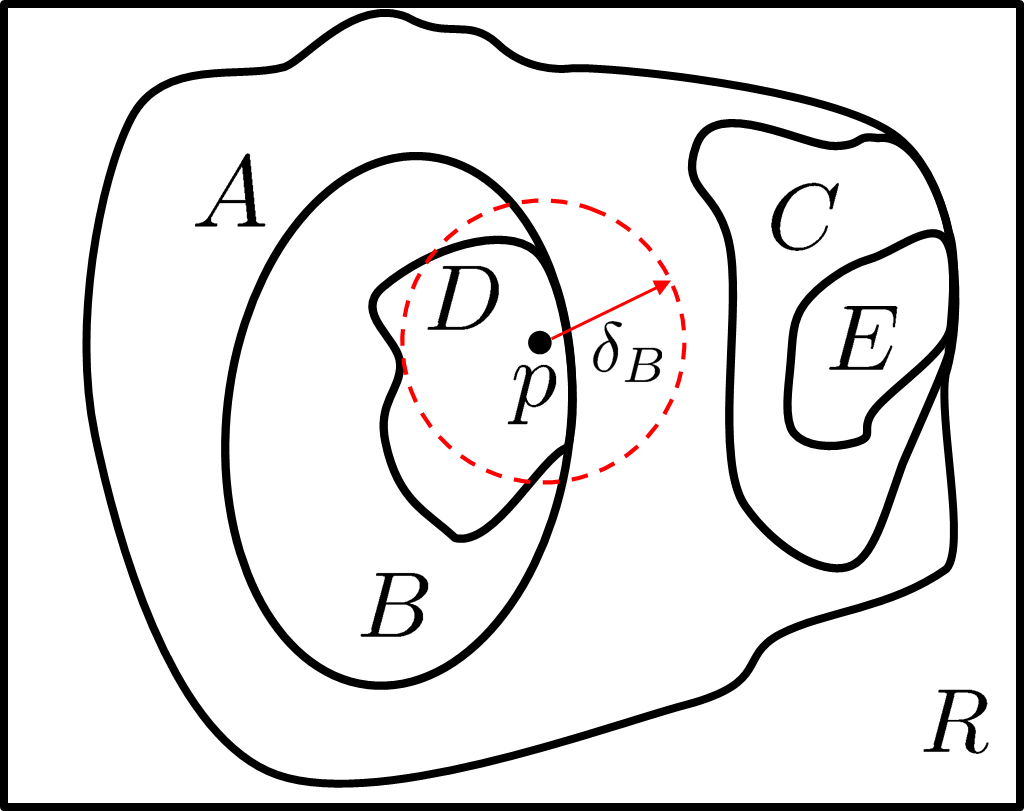}\\
(a) tree & (b) min-margin constraint at pixel $p$
\end{tabular}
\vspace{-2ex}
\end{center}
\caption{\labelstyle (a) tree $\Tree$ with 6 labels, $\Tree(B)$ is the subtree rooted at $B$ and $\parent(B)$ is $B$'s parent. (b) visually illustrates the min-margin constraint $\margins_B$ at an arbitrary pixel $p$ with label $\labelvar_p \in \Tree(B)$.
For a labeling $\labelvars$ to be valid w.r.t.~min-margin $\margins_B$; if $\labelvar_p \in \Tree(B)$ then any neighboring pixel $q$ within $\margins_B$ pixels from $p$ must be assigned to either $B$, one of its descendants or its parent, i.e.~$\labelvar_q \in \{\Tree(B)\cup \parent(B)\}$. Note that if $q$ was assigned to either one of $A$'s ancestors or $B$'s siblings this means we encountered the outside boundary of $A$ or $B$'s siblings within the $\margins_B$ margin.}
\label{fig:tconst}
\end{figure}

The {\em interaction} term in \eqref{eqn:OurMainEnergy} ensures that the min-margin  constraints are satisfied at every pixel, see Fig.~\ref{fig:tconst},
\begin{equation} \label{eqn:topologicalterm}
\TConst(\labelvars)=w_\infty
\sum_{\mathclap{\substack{\ell \in \labelset}}} \;\;
\sum_{\mathclap{\substack{p\in\pixelset \\ \labelvar_p \in \Tree(\ell)} }}
\quad\sum_{\substack{ q\in\pixelset\\ \lVert p-q\rVert < \margins_{\ell}}}\!\![\labelvar_q \not\in \{\Tree(\ell) \cup \parent(\ell)  \}]
\end{equation}
where $w_{\infty}$ is an infinitely large scalar, $\Tree(X)$ are the nodes of the subtree rooted at $X$, $\parent(X)$ is $X$'s parent in $\Tree$, and $[\;]$ is the {\em Iverson bracket}. This term guarantees that any labeling that violates min-margin constraint has infinite energy.

In general, the {\em interaction} term could model not only min-margin but also region attraction \cite{DB:ICCV09}, scene parsing \cite{olgalabelordering,DB:ICCV09}, or a combination of these constraints. However, the focus of this paper is developing an effective combinatorial optimization move for energy \eqref{eqn:OurMainEnergy}. Thus, for simplicity of exposition we only cover min-margins.

We now compare our formulation to that in \cite{DB:ICCV09}.
{\em Inclusion} is an easy constraint to impose in both formulations as it reduces to using tree-metric smoothness.
In our formulation {\em exclusion} is satisfied by definition because we use multi-label formulation and each pixel is assigned to only one label.
In contrast, in \cite{DB:ICCV09} the label of a pixel is represented by several binary variables.  Therefore, \cite{DB:ICCV09} needs to explicitly enforce exclusion to maintain the validity of these binary variables w.r.t.~tree $\Tree$. Often this leads to non-submodular terms that are difficult to optimize.

\begin{figure} [t]
\begin{center}
\begin{tabular}{@{\hspace{0ex}}c@{\hspace{1.0ex}}c}
\includegraphics[align=c,width=0.3\linewidth]{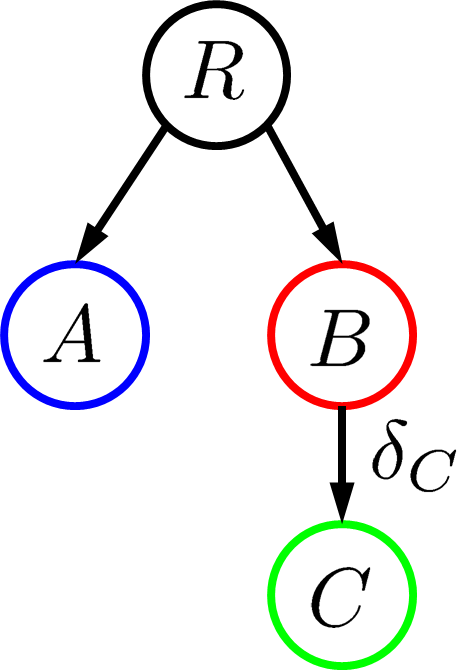}&
\includegraphics[align=c,width=0.45\linewidth]{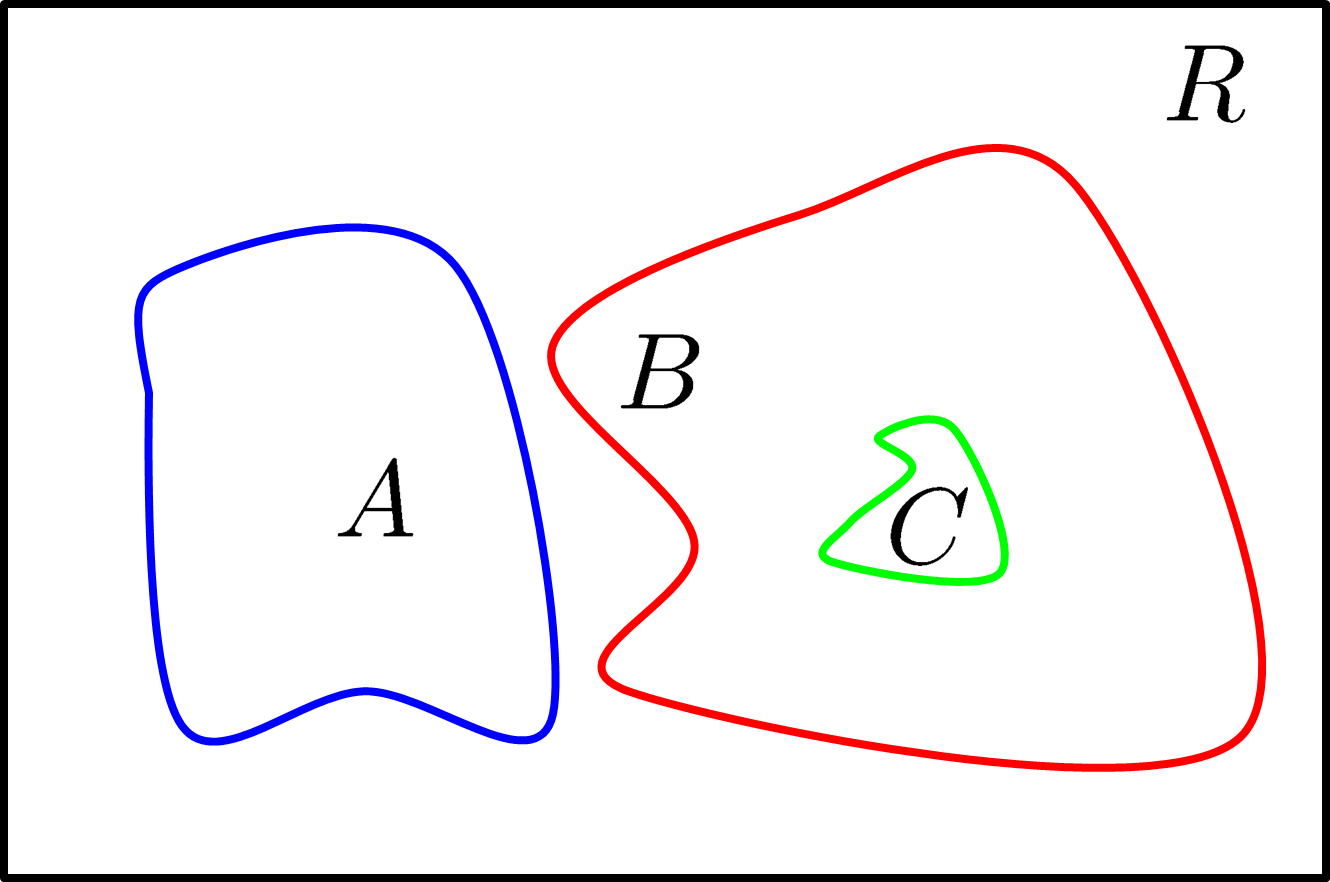}\\
(a) tree $\Tree$ \& margins $\margins$ & (b) current labeling\\\\
\includegraphics[align=c,width=0.45\linewidth]{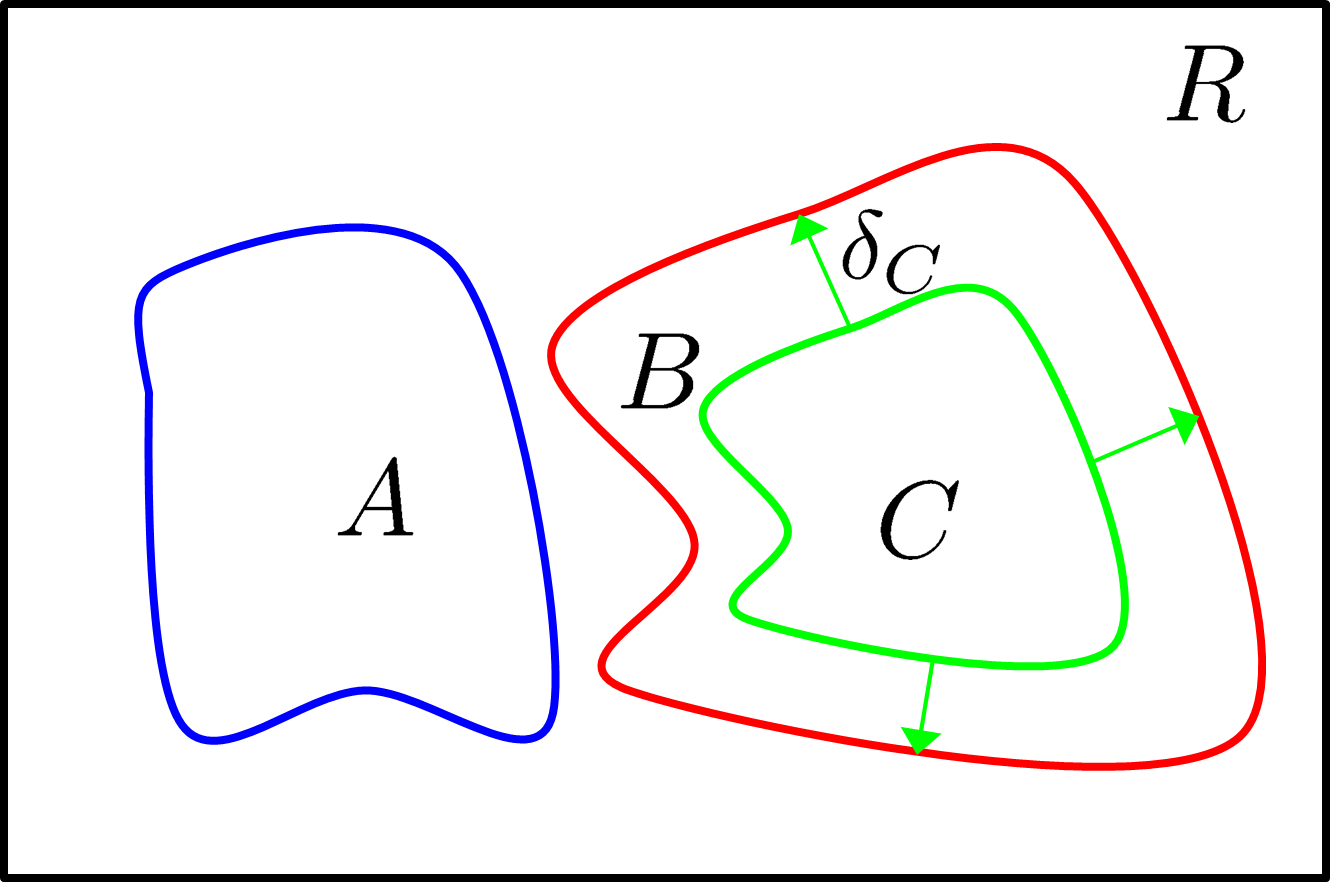}&
\includegraphics[align=c,width=0.45\linewidth]{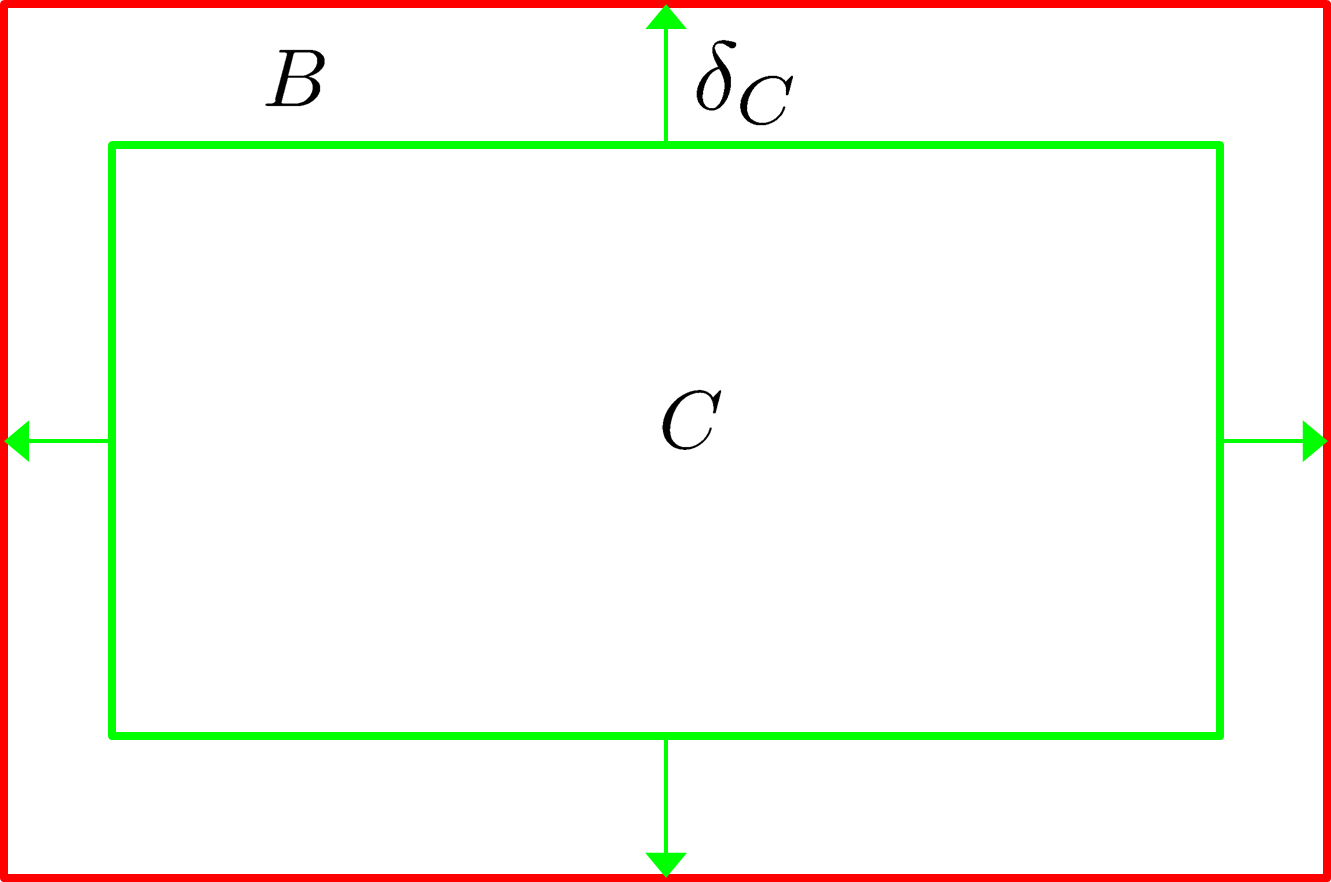}\\
(c) largest expansion \cite{BVZ:PAMI01} on $C$  & (d) largest Path-Move on $C$
\end{tabular}
\vspace{-2ex}
\end{center}
\caption{\labelstyle (a) shows tree and margins. (b) shows the current labeling. (c) and (d) show the largest possible expansion of label $C$ using binary expansion move \cite{BVZ:PAMI01} and our multi-label expansion move (Path-Move), respectively. Unlike \cite{BVZ:PAMI01}, Path-Move is capable of pushing all regions' boundaries when expanding $C$ without violating the interaction constraints.}
\label{fig:aexplocalminima}
\vspace{-3ex}
\end{figure}

\section{Optimization}
\label{sc:optimization}
In Section \ref{sc:pathmove} we introduce our Path-Move algorithm and in Section \ref{sc:pathmoverep} we show which interaction constraints Path-Move could optimize. The authors in \cite{DB:ICCV09} showed that HINTS is non-submodular for a general tree $\Tree$ and they used either QPBO or \a-exp for optimization. Unfortunately, QPBO does not guarantee to label all pixels and we observed that in our experiments, see Fig.~\ref{fig:teaser}. The \a-exp algorithm \cite{BVZ:PAMI01} is guaranteed to label all pixels but prone to weak local minima, Fig.~\ref{fig:teaser}.

We build on \a-exp algorithm \cite{BVZ:PAMI01}.
The algorithm in \cite{BVZ:PAMI01} maintains a valid current labeling $\clabelvars$ and iteratively tries to decrease the energy by switching from the current labeling to a nearby labeling via a {\em binary expansion} move. In a binary expansion, a label $\alpha \in \labelset$ is chosen randomly and allowed to expand. Each pixel is given a {\em binary} choice to either stay as $\clabelvar_p$ or switch to $\alpha$, i.e. $\labelvar_p\in \{\clabelvar_p,\alpha\}$. The algorithm stops when it cannot decrease the energy anymore.

\phantom{.}
\vspace{-4ex}
\begin{algorithm}
{\small
\DontPrintSemicolon
\hspace{0in}$\clabelvars := $ initial valid labeling\;
\hspace{0in}\textbf{repeat}\;
\hspace{.11in}   \textbf{for each} $\alpha \in \labelset$\;
\hspace{.22in}       $\labelvars^\alpha := \mathrm{arg} \min_{\labelvars} E(\labelvars\hspace{.01in})$ \,where $\labelvars$ is an $\a$-expansion of $\clabelvars\hspace{-.4in}$\;
\hspace{.22in}       \textbf{if} $E(\labelvars^\alpha) < E(\clabelvars\hspace{.008in})$ \;
\hspace{.33in}           $\clabelvars := \labelvars^\alpha$\;
\hspace{0in}\textbf{until} converged\;
}
\NoCaptionOfAlgo
\caption{{\small \textsc{A-Expansion Algorithm}~\cite{BVZ:PAMI01}\label{alg:alphaexpansion}}}
\end{algorithm}
\phantom{.}
\vspace{-5ex}

Due to the ``binary'' nature of the expansion move interaction constraints cause \a-exp to be highly sensitive to initialization and prone to converge to a weak local minima even for simple trees, see Fig.~\ref{fig:aexplocalminima}.

Instead of using a binary expansion move \cite{BVZ:PAMI01} in step 4 of the \a-exp algorithm, we propose a more powerful ``multi-label'' move, namely, Path-Move. Figure \ref{fig:aexplocalminima}(d) shows how robust a Path-Move is compared to a binary one \cite{BVZ:PAMI01}.

\subsection{Path-Move}
\label{sc:pathmove}
\begin{figure} [h]
\begin{center}
\begin{tabular}{cc}
\includegraphics[align=c,width=0.4\linewidth]{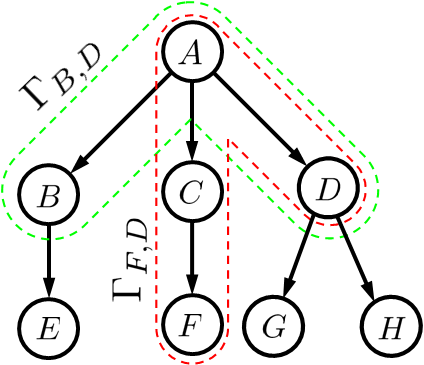}&
\includegraphics[align=c,width=0.4\linewidth]{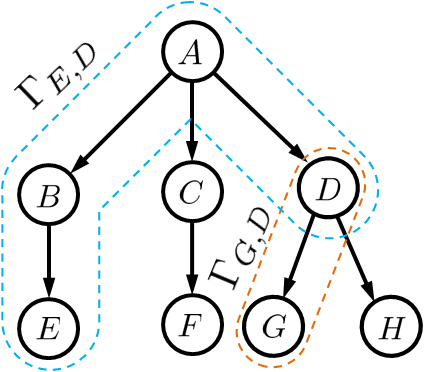}\\[-1ex]
\end{tabular}
\end{center}
\caption{\labelstyle shows for some $\Tree$ the sets of feasible labels when expanding on $D$ for pixels whose current labels are $B$ (green), $F$ (red), $E$ (blue) and $G$ (brown). Unlike Path-Move, in \cite{Ishikawa:PAMI03,olgarangemoves:IJCV12} the feasible set of labels during an expansion is the same for all pixels.}
\label{fig:gammaillustartion}
\vspace{-1ex}
\end{figure}

In a Path-Move on $\alpha$ each pixel $p$ can choose any label in the ordered set $\gpath[\clabelvar_p,\alpha]$ where $\clabelvar_p$ is the  current label of $p$.
Thus, the set of feasible labels for $p$ is $\gpath[\clabelvar_p,\alpha]$, see examples in Fig.~\ref{fig:gammaillustartion}.


Given an arbitrary $\Tree,$ current labeling $\clabelvars,$ and label $\alpha$, we now show how to build a graph such that the min-cut on this graph corresponds to the optimal Path-Move.
We use $s$ and $t$ to denote source and sink nodes of the min-cut problem, respectively. Our construction is motivated by \cite{Ishikawa:PAMI03,BVZ:PAMI01,olgarangemoves:IJCV12}.

\begin{figure}[t]
\vspace{-2ex}
\begin{center}
\begin{tabular}{@{\hspace{-1.5ex}}c@{\hspace{1ex}}c}
\includegraphics[align=c,height=0.32\textheight]{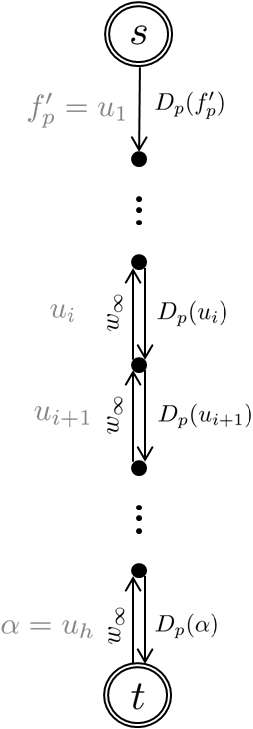}&
\includegraphics[align=c,height=0.32\textheight]{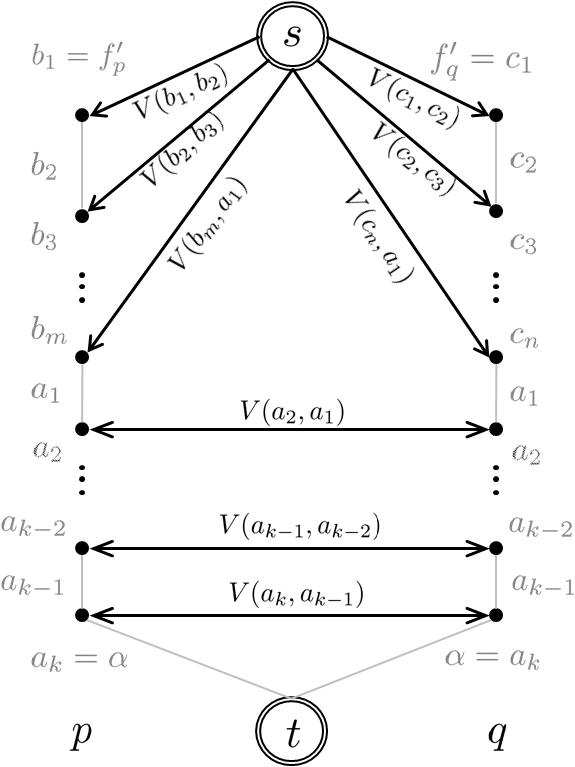}\\
{\small (a) data encoding} &
{\small (b) smoothness encoding}\\[-1ex]
\end{tabular}
\end{center}
\caption{{\labelstyle (a) shows the part of our graph that encodes the data term of pixel $p$. The black nodes represent ${\cal C}_p$.
Next to each edge along $(s,{\cal C}_p,t)$ we show in light grey the label that pixel $p$ is assigned to if that edge is cut.
(b) shows the part of our graph that encodes the smoothness term for neighboring pixels $p$ and $q$ with current labels $\clabelvar_p$ and $\clabelvar_q$, respectively.
Grey edges in (b) are those ones illustrated in (a) but redrawn in (b) without their weights for clarity.}}
\label{fig:pathmoveconstruction}
\vspace{-1ex}
\end{figure}

\paragraph{Data Term:}
For each pixel $p$ we generate a chain of nodes ${\cal C}_p$ whose size is $|\gpath[\clabelvar_p,\alpha]|-1$.
Let us rename $\gpath[\clabelvar_p,\alpha]$ to $(u_1, u_2,\hdots, u_h)$ where $u_1=\clabelvar_p$ and $u_h=\alpha$. Note that $u_i$ and $h$ depend on $p$ but we drop explicit dependence on $p$ from notation for clarity. Figure \ref{fig:pathmoveconstruction}(a) illustrates chain ${\cal C}_p$ and how it is linked to $s$ and $t$.
The edge weights along the directed path $(s,{\cal C}_p,t)$ encode the the data terms of $p$ while the weights along the opposite direction are $w_\infty$. If the $i^{th}$ edge along the $(s,{\cal C}_p,t)$ is cut, then pixel $p$ is assigned to label $u_i$. The $w_\infty$ edges ensure that any min-cut severs only one edge on the $(s,{\cal C}_p,t)$ path as proposed by \cite{Ishikawa:PAMI03}. Thus, the sum of severed edges on paths $(s,{\cal C}_p,t)$ for all pixels $p \in \pixelset$ adds to the data term in \eqref{eqn:OurMainEnergy}.

It should be noted that although $\unary_p$ is used as a weight for n-link\footnote{An edge between two nodes and neither of those nodes is $s$ or $t$.} edges it could still be negative. In case of negative $\unary_p$ a positive constant $K$ is added to $\unary_p(\ell)$ for all $p\in \pixelset$ and $\ell \in \labelset$ to ensure that the new data terms are non-negative for every pixel---equivalent to adding a constant $|\pixelset|K$ to \eqref{eqn:OurMainEnergy}.

\paragraph{Smoothness:}
Let $p$ and $q$ be a pair of neighboring pixels. Note the overlap between $\gpath[\clabelvar_p,\alpha]$ and $\gpath[\clabelvar_q,\alpha]$ is at least one label, see Fig.~\ref{fig:gammaillustartion}. Our graph construction treats the sequence of overlapping labels of paths $\gpath[\clabelvar_p,\alpha]$ and $\gpath[\clabelvar_q,\alpha]$ differently from the non-overlapping parts. Therefore we rename $\gpath[\clabelvar_p,\alpha]=(b_1,\hdots,b_m,a_1,\hdots,a_k)$ and $\gpath[\clabelvar_q,\alpha] =(c_1,\hdots,c_n,a_1,\hdots,a_k)$ to emphasize the overlap. Figure~\ref{fig:pathmoveconstruction}(b) shows the newly weighted edges that are added to out constructed graph to encode the smoothness penalty $\pairwise_{pq}$.

The overlapping part $(a_1,\hdots,a_k)$ forms a linear ordering for which the smoothness cost is encoded as proposed by \cite{Ishikawa:PAMI03}. The non-overlapping parts $(b_1,\hdots,b_m,a_1)$ and $(c_1,\hdots,c_m,a_1)$ each forms a linear ordering independent of the other, but extending $(a_1,\hdots,a_k)$ linear ordering.
In this case smoothness penalties are handled by additional edges from $s$, for proof of correctness see Appendix \ref{ap:smoothnessproof}.

\vspace{-1ex}
\paragraph{Interaction Constraints:} Let $p$ and $q$ be $\margins_A>0$ within each other. As per energy \eqref{eqn:topologicalterm}, to impose the $\margins_A$ margin constraint between $p$ and $q$ we need to add edges to our graph to ensure that whenever $\labelvar_p \in \Tree(A)$ and $\labelvar_q \not \in \{\Tree(A) \cup \parent(A)\}$ the corresponding energy is infinite.
Thus, we need to eliminate/forbid labels along the $\gpath[\clabelvar_q,\alpha]$ path that would violate the $\margins_A$ constraints if $p$ is assigned to a label in $\Tree(A).$
To impose such constraint between $p$ and $q$ expansion paths there are several cases to consider depending on whether each of $\alpha$, $\clabelvar_p$ or $\clabelvar_q$ is in $\Tree(A)$ or not as follows.

\begin{figure}[h]
\begin{center}
\begin{tabular}{@{\hspace{-1ex}}c@{\hspace{1ex}}c}
\includegraphics[align=c,width=0.43\linewidth]{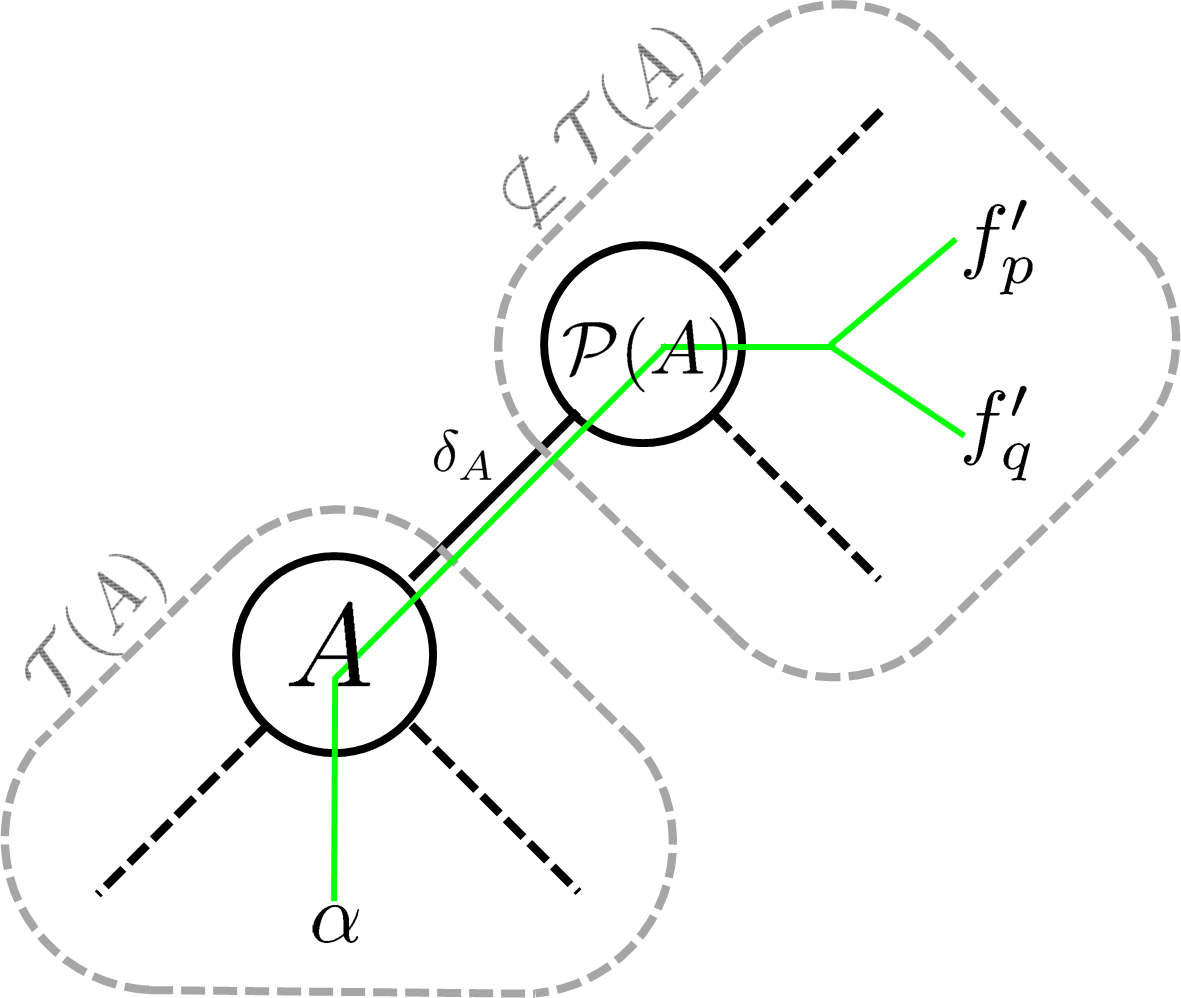}&
\includegraphics[align=c,width=0.43\linewidth]{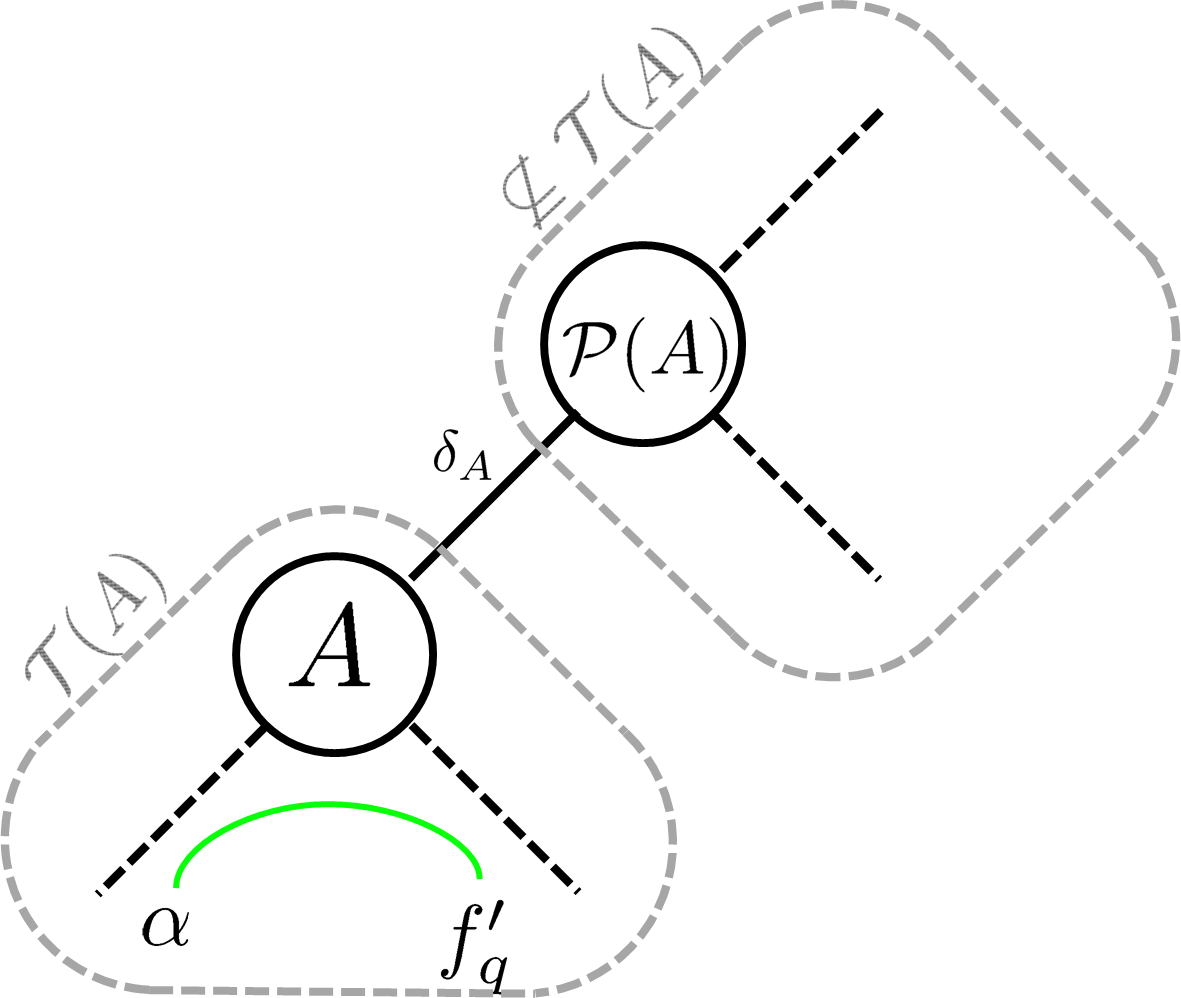}\\
{\small (a) Scenario I, case 1} &
{\small (b) Scenario I, case 2}
\end{tabular}
\end{center}
\caption{{\labelstyle (a) shows that for Scenario I case 1 any possible $p$ or $q$ expansion paths (shown in green) must include $A$ and $\parent(A)$. (b) shows that for Scenario I case 2 the expansion path $\gpath[\clabelvar_q,\a]$ (shown in green) must be fully contained in $\Tree(A)$. If the aforementioned deductions are invalid then the undirected $\Tree$ is not an acyclic graph, and this violates our assumption that $\Tree$ is a tree.}}
\label{fig:interactioncase1and2}
\end{figure}

\begin{figure}[h]
\begin{center}
\begin{tabular}{@{\hspace{-2ex}}c@{\hspace{7ex}}c}
\includegraphics[align=c,width=0.43\linewidth]{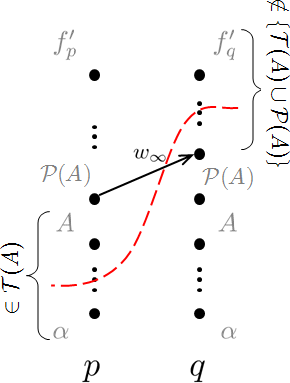}&
\includegraphics[align=c,width=0.43\linewidth]{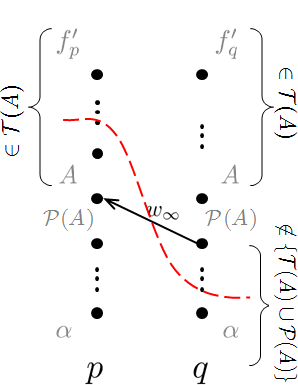}\\\\[-1ex]
{\small (a) $\alpha \in \Tree(A)$ } &
{\small (b) $\alpha \not \in\Tree(A)$}\\
{\small \phantom{6ex}$\quad\quad\clabelvar_p, \clabelvar_q \not \in \Tree(A)$ } &
{\small \phantom{6ex}$\quad\quad\clabelvar_p,\clabelvar_q \in \Tree(A)$}\\[-1ex]
\end{tabular}
\end{center}
\caption{ {\labelstyle (a) and (b) show the required $w_\infty$ edge for the two main cases that occur when imposing the $\margins_A$ margin constraint. The red dashed curves illustrate prohibitively expensive cuts that violate $\margins_A$ margin constraint.}}
\label{fig:pathmoveminmargin}
\end{figure}

\noindent\textbf{Scenario I} when $\alpha \in \Tree(A)$:

\vspace{0.5ex}
Case 1, assume $\clabelvar_p \not \in \Tree(A)$ and $\clabelvar_q \not \in \Tree(A)$. Since $\clabelvar_p \not \in \Tree(A)$ and $\alpha \in \Tree(A)$ by assumption, we can deduce that $A$ and $\parent(A)$ are both in $\gpath[\clabelvar_p,\alpha]$ see Fig.\ref{fig:interactioncase1and2} (a), otherwise our assumption that $\Tree$ is a tree would be violated. Following the same reasoning we can deduce that  $A$ and $\parent(A)$ are both in $\gpath[\clabelvar_q,\alpha]$.
Thus, there are possible labelings/configurations involving $p$ and $q$ that violate $\margins_A$.
In other words, if $p$ is assigned to a label in $\Tree(A)$ there is a chance of assigning $q$ to a label that is not in $\{\Tree(A) \cup \parent(A)\}$, since part of $q$ expansion path is not entirely in $\{\Tree(A) \cup \parent(A)\}$, Fig.~\ref{fig:interactioncase1and2}(a).
We forbid those configurations by adding a $w_\infty$ edge between graph chains ${\cal C}_p$ and ${\cal C}_q$ as shown in Fig.\ref{fig:pathmoveminmargin}(a). Thus, eliminating the possibility of a min.~cut that would simultaneously assign $q$ to a label $ \labelvar_q \not \in \{\Tree(A) \cup \parent(A)\}$ and $p$ to a label $\labelvar_p \in \Tree(A).$

\vspace{0.5ex}
Case 2, assume $\clabelvar_q \in \Tree(A)$. Since $\clabelvar_q \in \Tree(A)$ and $\alpha \in \Tree(A)$ by assumption, we can deduce that $\gpath[\clabelvar_q,\alpha] \subseteq \Tree(A)$, otherwise our assumption that $\Tree$ is a tree would be violated, see Fig.~\ref{fig:interactioncase1and2} (b).
Thus, additional edges are not needed since $\labelvar_q$ is guaranteed to be in $\Tree(A)$. Simply, in this case there is no chance of violating the $\margins_A$ constraints regardless of what label is assigned to $p$, because we could only assign $q$ to a label in $\Tree(A)$, i.e.~$\labelvar_q \in \{\Tree(A) \cup \parent(A)\}$.

\vspace{0.5ex}
Case 3, assume $\clabelvar_p \in \Tree(A)$ and $\clabelvar_q \not \in \Tree(A)$. Since $\alpha \in \Tree(A)$ and $\clabelvar_q\not\in \Tree(A)$ by assumption, we can deduce that $\parent(A)\in \gpath[\clabelvar_q,\alpha]$, otherwise $\Tree$ is not a tree.
On the one hand, if $\clabelvar_q =\parent(A)$ then no additional edges needed since $\labelvar_q \in \{ \parent(A) \cap \Tree(A)\}$. On the other hand, the case when $\clabelvar_q \not= \parent(A)$ is not possible as this would imply that the current labeling violates the $\margins_A$. Recall, Path-Moves switches from one valid labeling to another.

\vspace{0.5ex}
\noindent\textbf{Scenario II} when $\alpha \not \in \Tree(A)$:

\vspace{0.5ex}
Case 1, assume $\clabelvar_p \in \Tree(A)$ and $\clabelvar_q \in \Tree(A)$. This case follows the same reasoning as scenario I, case 1. Similarly to scenario I, case 1 we handle the forbidden configurations by adding an additional $w_\infty$ edge as shown in Fig.\ref{fig:pathmoveminmargin}(b).

\vspace{0.5ex}
Case 2, assume $\clabelvar_p  \not  \in \Tree(A)$. Since $\clabelvar_p$ and $\alpha$ are $\not \in \Tree(A)$ by assumption, we can deduce that $\labelvar_p \not \in \Tree(A)$. Thus, no new edges are needed, as we are only interested in the case when $\labelvar_p \in \Tree(A)$. Recall that we only forbid labels along chain ${\mathcal C}_q$  if $\labelvar_p$ is in $\Tree(A)$, which is not possible in this case.

\vspace{0.5ex}
Case 3, assume $\clabelvar_p \in \Tree(A)$ and $\clabelvar_q \not \in \Tree(A)$.
Since $\clabelvar_q$ and $\alpha$ are  $\not \in \Tree(A)$ by assumption then we can deduce that $\gpath[\clabelvar_q,\alpha] \not \subseteq \Tree(A).$
On the one hand, if $\parent(A) \not \in \gpath[\clabelvar_q,\alpha]$ then we can deduce that $\clabelvar_q \not \in \{\Tree(A)\cup \parent(A)\}$ while $\clabelvar_p \in \Tree(A)$ (by assumption), i.e.~current labeling violates the $\margins_A$ constraint and this is not possible.
On the other hand, if $\parent(A) \in \gpath[\clabelvar_q,\a]$ then we can deduce that $\clabelvar_q = \parent(A)$ otherwise the current labeling would violate $\margins_A$. When $\clabelvar_q=\parent(A)$ the construction is as shown in Fig.\ref{fig:pathmoveminmargin}(b) except there are no nodes above $\parent(A)$ for $q$.

\subsection{Interaction Representability Condition}
\label{sc:pathmoverep}
Unfortunately, not every interaction constraint could be represented as constraints between expansion paths during a Path-Move. To be specific, sometimes interactions lead to conflicting constraints between the expansion paths, i.e.~the graph construction inadvertently forbids a permissible configuration. We refer to an interaction that could be optimized by Path-Move as a Path-Move representable constraint, e.g.~min-margin is a Path-Move representable constraint while strict box-layout constraints described in \cite{olgalabelordering,DB:ICCV09} for scene parsing is not Path-Move representable.

The objective of box-layout scene parsing is segmenting the scene into 5 regions; {\em left}, {\em top}, {\em right}, {\em bottom}, and {\em background}, denoted by L, T, R, B, and G, respectively, in Fig.~\ref{fig:strictboxlayout}(a). The strict box-layout constraints are illustrated in Fig.~\ref{fig:strictboxlayout}(b-f). For example as shown in (b), when pixel $q$ is directly above $p$ and $f_p=G$ then according to the layout in (a) $q$ could only be labeled $G$ or $T$, i.e.~$f_q \not \in \{L,R,B\}$. The rest of the constraints shown in Fig.~\ref{fig:strictboxlayout}(b-f) are derived in the same way from the layout in (a).

\newcommand{\boxwidth}{0.35\linewidth}
\newcommand{\boxinspace}{\hspace{6ex}}
\begin{figure}
\begin{center}
\begin{tabular}{c@{\boxinspace}c}
\includegraphics[align=c,width=\boxwidth]{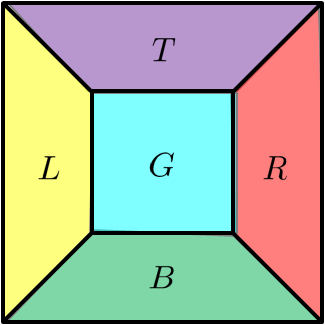}&
\includegraphics[align=c,width=\boxwidth]{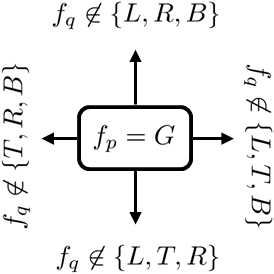}\\
{\small (a) strict box-layout} & {\small (b) G restrictions}\\\\
\includegraphics[align=c,width=\boxwidth]{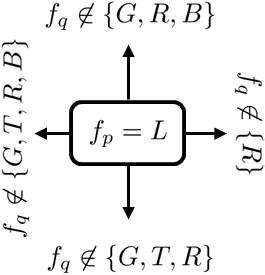}&
\includegraphics[align=c,width=\boxwidth]{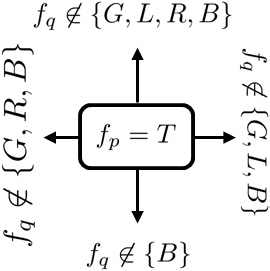}\\
{\small (c) L restrictions} & {\small (d) T restrictions}\\\\
\includegraphics[align=c,width=\boxwidth]{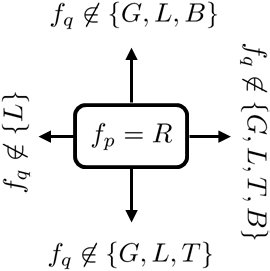}&
\includegraphics[align=c,width=\boxwidth]{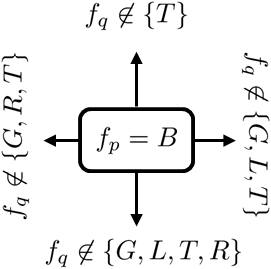}\\
{\small (e) R restrictions} & {\small (f) B restrictions}
\end{tabular}
\end{center}
\caption{{\labelstyle (a) shows a strict box-layout configuration. (b-f) show the strict box-layout constraints corresponding. Note that those constraints are direction dependent.}}
\label{fig:strictboxlayout}
\end{figure}

\begin{figure}
\vspace{-1ex}
\begin{center}
\begin{tabular}{cc}
\includegraphics[align=c,width=0.43\linewidth]{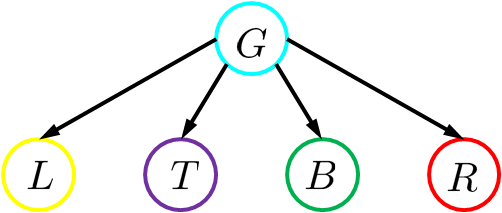}&
\includegraphics[align=c,width=0.43\linewidth]{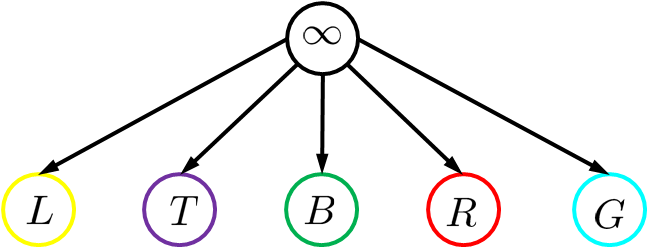}\\
{\small (a) tree 1} & {\small (b) tree 2}\\[-1.5ex]
\end{tabular}
\end{center}
\caption{{\labelstyle show two possible hierarchical trees. The $\infty$ label is an artificial label with $\infty$ data cost, i.e.~no pixel could be assigned to it. The strict box-layout constraints are Path-Move representable for tree 2 but not tree 1.}}
\label{fig:boxlayouttrees}
\vspace{-1ex}
\end{figure}

\begin{figure}
\begin{center}
\begin{tabular}{@{\hspace{0ex}}c@{\hspace{5ex}}c}
\includegraphics[align=c,width=0.55\linewidth]{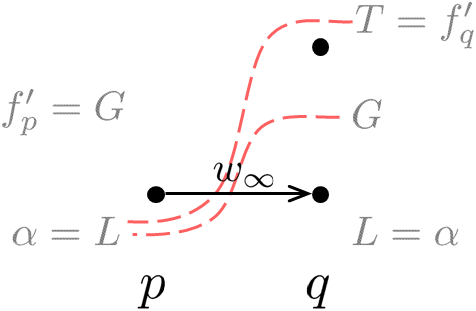}&
\includegraphics[align=c,width=0.32\linewidth]{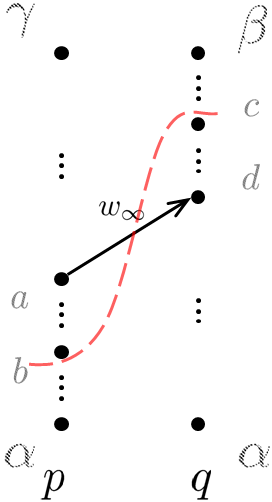}\\[-1ex]
\end{tabular}
\end{center}
\caption{{\labelstyle (Left) shows a case of conflicting interaction constraints between $p$ and $q$ expansion paths. In this setting the $p$ and $q$ labeling $[\labelvar_p,\labelvar_q]$ is allowed to be $[L,T]$ but not $[L,G]$. However, by forbidding $[L,G]$ we also forbid $[L,T]$. (Right) in general, a Path-Move can not simultaneously permit configuration $[b,c]$ while prohibiting $[a,d]$. The $w_\infty$ edge that forbids [a,d] also forbids $[b,c]$. The red curves are forbidden min.~cuts.}}
\label{fig:pathmoverepsentable}
\vspace{-3ex}
\end{figure}

\newpage To show a case that leads to conflicting constraints let us consider the tree shown in Fig.~\ref{fig:boxlayouttrees}(a). Also assume that pixel $p$ is directly below $q$ and that their current labeling is $G$ and $T$, respectively.
As shown in Fig.~\ref{fig:pathmoverepsentable}(Left), when expanding on $L$ an $w_\infty$ edge is add to avoid assigning $p$ and $q$ to $L$ and $G$, respectively, which is a forbidden configuration as per Fig.~\ref{fig:strictboxlayout}(c). However, the same edge also forbids a permissible labeling that would assign $p$ and $q$ to $L$ and $T$, respectively.
As you can see, when using the hierarchical tree in Fig.~\ref{fig:boxlayouttrees}(a) we can not properly represent the strict box-layout interaction constraints during a Path-Move.

\begin{figure}
\begin{center}
\begin{tabular}{c@{\boxinspace}c}
\includegraphics[align=c,width=\boxwidth]{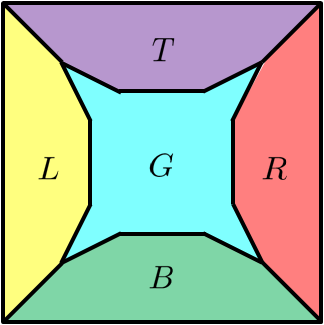}&
\includegraphics[align=c,width=\boxwidth]{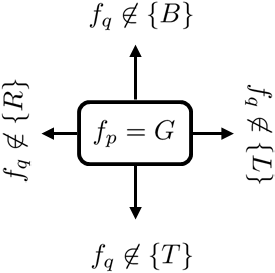}\\
{\small (a) relaxed box-layout} & {\small (b) G restrictions}\\\\
\includegraphics[align=c,width=\boxwidth]{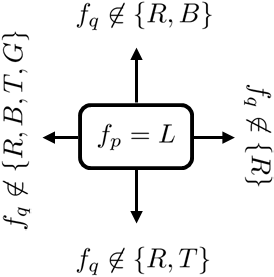}&
\includegraphics[align=c,width=\boxwidth]{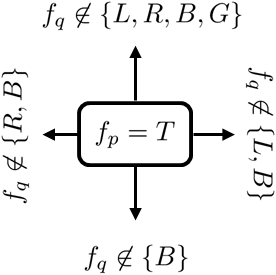}\\
{\small (c) L restrictions} & {\small (d) T restrictions}\\\\
\includegraphics[align=c,width=\boxwidth]{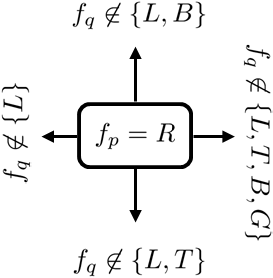}&
\includegraphics[align=c,width=\boxwidth]{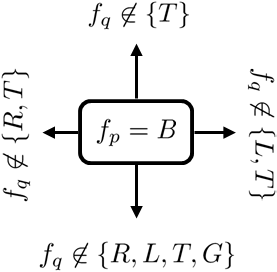}\\
{\small (e) R restrictions} & {\small (f) B restrictions}
\end{tabular}
\end{center}
\caption{{\labelstyle (a) shows the relaxed box-layout configuration. (b-f) show the direction dependent constraints corresponding to the relaxed box-layout.}}
\label{fig:relaxedboxlayout}
\end{figure}

In general, an interaction constraint is not Path-Move representable if there exists $\alpha, \beta$ and $\gamma \in \labelset$ where $a < b  \in \gpath[\gamma,\alpha]$ and $c < d \in \gpath[\beta,\alpha]$ while configuration $[a,d]$ is prohibited and $[b,c]$ is permissible, see Fig.~\ref{fig:pathmoverepsentable} (Right). Nonrepresentable Path-Move interactions lead to a nonsubmodular Path-Move \cite{schlesinger2006transforming}. Nonetheless, this could be avoided either by modifying the tree or relaxing the interaction constraints. For instance, the strict box-layout becomes Path-Move representable when using the alternative hierarchical tree shown in Fig.~\ref{fig:boxlayouttrees}(b).
If modifying the hierarchical tree is not an option, then by relaxing the constraints one could always achieve Path-Move representability. For example, by relaxing the strict box-layout constraints shown in Fig.~\ref{fig:relaxedboxlayout} they become Path-Move representable for the hierarchical tree shown in Fig.~\ref{fig:boxlayouttrees}(a).

\section{Shape Priors for HINTS}
\label{sc:shapeprior}
In this section we extend star-shape \cite{olga:ECCV08}, Geodesic-star \cite{gulshan2010geodesic} and Hedgehogs \cite{isackhedgehog} priors to the HINTS model and show how to enforce these priors during a Path-Move.

\begin{figure}[t]
\begin{center}
\begin{tabular}{cc}
\includegraphics[align=c,width=0.45\linewidth]{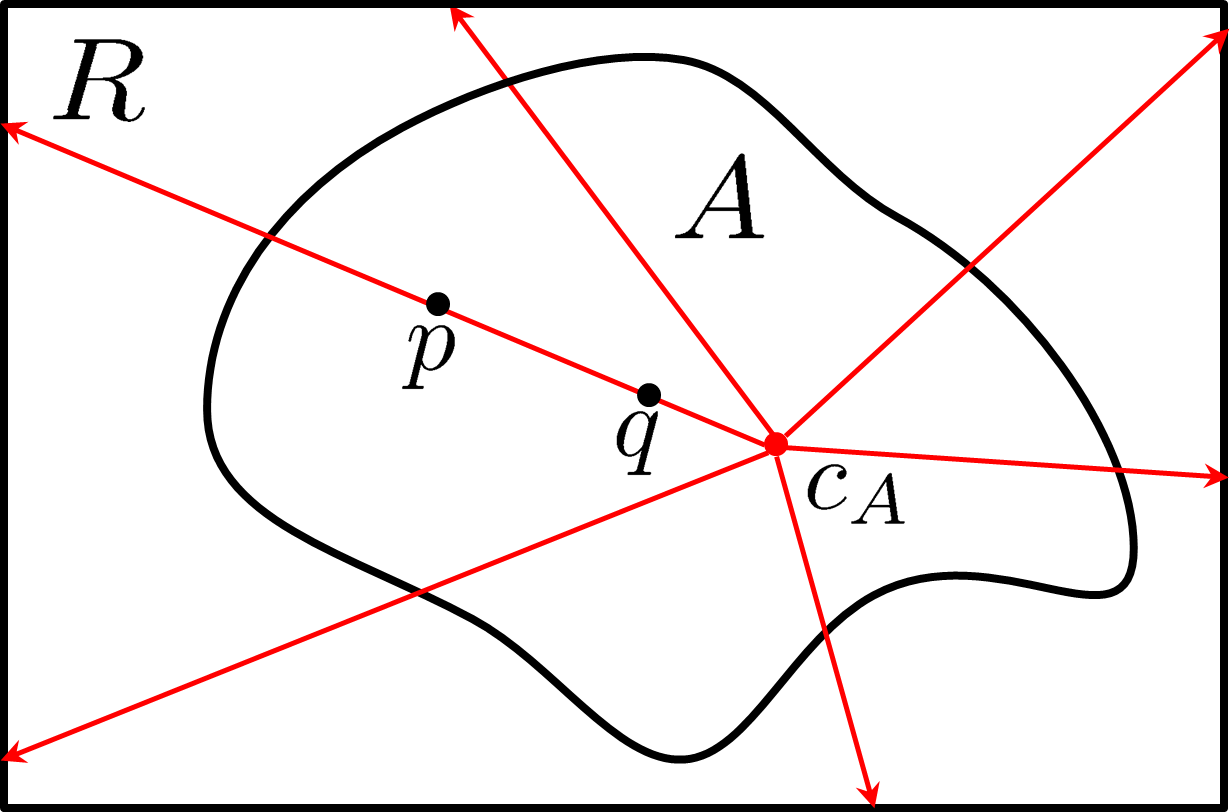}&
\includegraphics[align=c,width=0.45\linewidth]{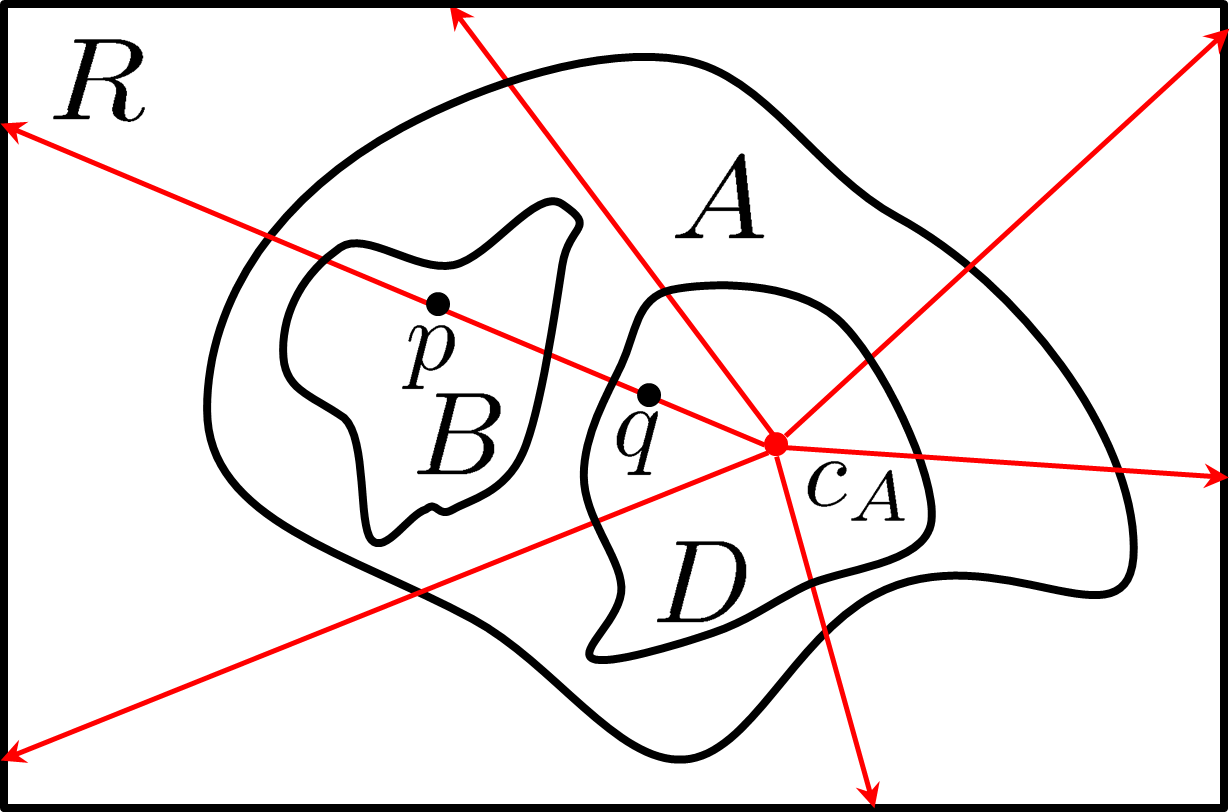}\\
{\small (a) star-shape prior \cite{olga:ECCV08} } &
{\small (b) star-shape prior + HINTS}
\end{tabular}
\end{center}
\caption{ {\labelstyle  (a) and (b) illustrate star-shape prior constraint for label $A$ in binary and partially ordered segmentations, respectively. The star-center is denoted by $c_A$. Both (a) and (b) show a valid star-shape for label A.}}
\label{fig:starshapedef}
\end{figure}

In the context of binary segmentation, star-shape prior \cite{olga:ECCV08} on label $A$ with star center $c_A$ reduces to the following constraint.
If  pixels $p$ and $q$ lie on any line originating from $c_A$ with $q$ in the middle and $p$ is labeled $A$, then $q$ must also be labeled $A$, see Fig.\ref{fig:starshapedef}(a).
Geodesic-star \cite{gulshan2010geodesic} and Hedgehogs \cite{isackhedgehog} differ from star-shape prior in terms of what defines the center and how  lines from the center (or geodesic paths) are generated. Furthermore, Hedgehogs \cite{isackhedgehog} allow control over shape constraint tightness, see \cite{isackhedgehog} for details.

For partially ordered segments we generalize the star-shape prior constraint as follows. If  pixels $p$ and $q$ lie on any line originating from $c_A$ with $q$ in the middle and $f_p$ is in ${\mathcal T}(A)$, then $f_q$ must also be in ${\mathcal T}(A)$, see Fig.\ref{fig:starshapedef}(b).

The shape prior penalty term is
\begin{equation} \label{eq:shapprior}
S(\labelvars)=w_{\infty}\sum_{\ell \in \labelset}\quad \sum_{\mathclap{\substack{p \in \pixelset\\ \labelvar_p \in \Tree(\ell)}}}\quad \sum_{pq \in {\mathcal S}_{\ell}} [\labelvar_q \not \in \Tree(\ell)],
\end{equation}
where ${\mathcal S}_\ell$ is the set of all ordered pixel pairs\footnote{In practice, it is enough to include only consecutive pixel pairs in ${\cal S}_\ell$.} $(p,q)$ along any line containing $c_\ell$ such that $q$ is between $p$ and $c_\ell$. Using \cite{gulshan2010geodesic} or \cite{isackhedgehog} instead of \cite{olga:ECCV08} results in a different ${\cal S}_\ell$.

\begin{figure}[h]
\begin{center}
\begin{tabular}{c@{\hspace{6ex}}c}
\includegraphics[align=c,width=0.43\linewidth]{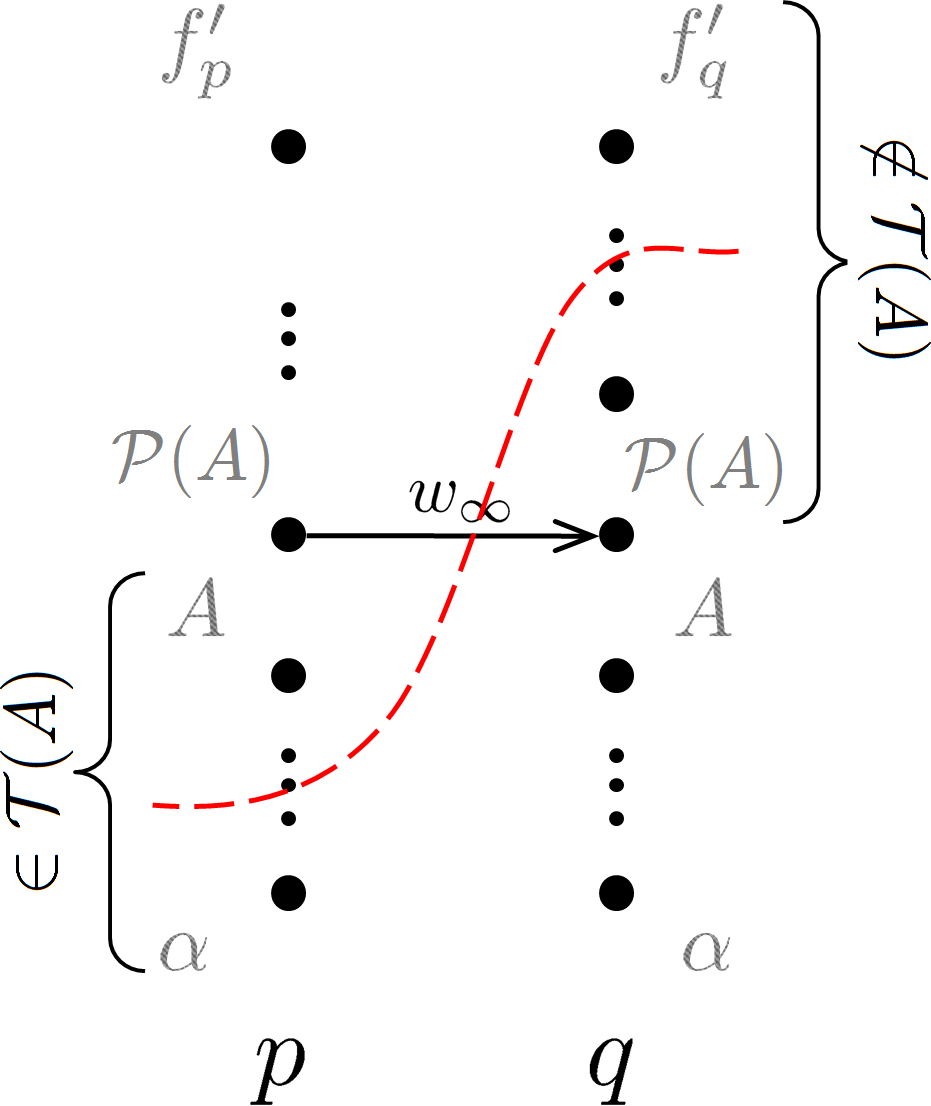}&
\includegraphics[align=c,width=0.43\linewidth]{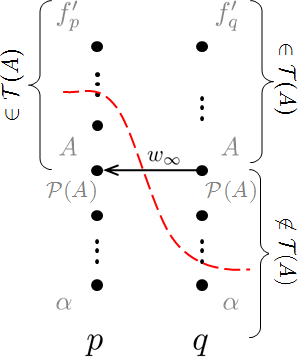}\\\\
{\small (a) $\alpha \in \Tree(A)$ } &
{\small (b) $\alpha \not \in\Tree(A)$}\\
{\small \phantom{6ex}$\quad\quad\clabelvar_p, \clabelvar_q \not \in \Tree(A)$ } &
{\small \phantom{6ex}$\quad\quad\clabelvar_p,\clabelvar_q \in \Tree(A)$}\\
\end{tabular}
\end{center}
\caption{ {\labelstyle Assume pixels $p$ and $q$ lie on a line originating from star-center $c_A$ of label $A$, and that $q$ lies between $c_A$ and $p$. (a) and (b) show the two cases that require a $w_\infty$ edge to impose the star-shape prior on Label $A$. The red dashed curves are prohibitively expensive cuts that violate the star-shape constraint.}}
\label{fig:pathmoveshapepriorconst}
\end{figure}

Let pixels $p$ and $q$ lie on a line passing through $c_A$, and $q$ is between $c_A$ and $p$. To impose the star-shape prior for label $A$ during a Path-Move, there are multiple cases to consider depending on whether each of $\alpha$, $\clabelvar_p$ and $\clabelvar_q$ is in $\Tree(A)$ or not as follows.

\noindent\textbf{Scenario I:} when $\alpha \in \Tree(A)$:

Case 1, assume $\clabelvar_p \not \in \Tree(A)$ and $\clabelvar_q \not \in \Tree(A)$. In this case we can deduce that $A$ and $\parent(A)$ are both in $\gpath[\clabelvar_p,\alpha]$ and $\gpath[\clabelvar_q,\alpha]$. Thus, there are possible forbidden configurations and to handle them we add an $w_\infty$ edge as in Fig.~\ref{fig:pathmoveshapepriorconst}(a).

Cases 2, assume $\clabelvar_q \in \Tree(A)$.
We can deduce that $\gpath[\clabelvar_q,\alpha] \subseteq \Tree(A)$. Thus, no additional edges are needed since $\labelvar_q$ is guaranteed to be in $\Tree(A)$.

Case 3, assume $\clabelvar_p \in \Tree(A)$ and $\clabelvar_q \not \in \Tree(A)$. An impossible case as the current labeling would be violating the shape-prior.
\begin{table*}[t]
\begin{center}
\begin{tabular}{l|*{12}{c|}}
\cline{2-13}
& \multicolumn{3}{c|}{Grey-Matter}
& \multicolumn{3}{c|}{White-Matter}
& \multicolumn{3}{c|}{CSF}
& \multicolumn{3}{c|}{SGM}\\
\cline{2-13}
& Ours & QPBO  & \a-exp & Ours & QPBO & \a-exp & Ours & QPBO & \a-exp & Ours & QPBO & \a-exp \\
\hline
\multicolumn{1}{|c|}{F$_1$ Score}&$\mathbf{0.92}$ & $0.83$ & $0.32$ & $\mathbf{0.92}$ & $0.90$ & $0.56$ & $\mathbf{0.85}$ & $0.82$ & $0.04$ & $\mathbf{0.83}$ & $0.81$ & $0.37$\\
\hline
\multicolumn{1}{|c|}{Precision}&$0.87$ & $0.87$ & $0.88$ & $0.92$ & $0.92$ & $0.46$ & $0.78$ & $0.83$ & $0.02$ & $0.92$ & $0.93$ & $0.23$\\
\hline
\multicolumn{1}{|c|}{Recall} &$0.97$ & $0.80$ & $0.19$ & $0.93$ & $0.88$ & $0.74$ & $0.93$ & $0.82$ & $0.56$ & $0.76$ & $0.71$ & $0.92$\\
\hline
\end{tabular}
\end{center}
\caption{{\labelstyle Compares our Path-Moves optimization to QPBO \cite{rotherQPBO} and \a-exp \cite{BVZ:PAMI01} which were proposed by \cite{DB:ICCV09}. The precision and recall were averaged over 15 examples. Our method and QPBO clearly outperformed \a-exp which was very sensitive to initialization and the order in which labels were expanded on.
On average QPBO left $2.8\%$ of the pixels unlabeled and in one instance $7\%$. These values rise significantly when not using the Hedgehog shape prior, see Table \ref{tbl:brainnohhog}.}}
\label{tbl:brainhhog}
\vspace{-1ex}
\end{table*}

\begin{table*}
\begin{center}
\begin{tabular}{l|*{12}{c|}}
\cline{2-13}
& \multicolumn{3}{c|}{Grey-Matter}
& \multicolumn{3}{c|}{White-Matter}
& \multicolumn{3}{c|}{CSF}
& \multicolumn{3}{c|}{SGM}\\
\cline{2-13}
& Ours & QPBO  & \a-exp & Ours & QPBO & \a-exp & Ours & QPBO & \a-exp & Ours & QPBO & \a-exp \\
\hline
\multicolumn{1}{|c|}{F$_1$ Score}&$\mathbf{0.92}$ & $0.53$ & $X$ & $\mathbf{0.93}$ & $0.90$ & $0.58$ & $\mathbf{0.84}$ & $0.77$ & $0.05$ & $\mathbf{0.84}$ & $0.79$ & $0.02$\\
\hline
\multicolumn{1}{|c|}{Precision}&$0.87$ & $0.82$ & $X$ & $0.93$ & $0.95$ & $0.47$ & $0.77$ & $0.84$ & $0.03$ & $0.90$ & $0.90$ & $0.11$\\
\hline
\multicolumn{1}{|c|}{Recall} &$0.97$ & $0.39$ & $X$ & $0.93$ & $0.85$ & $0.76$ & $0.92$ & $0.71$ & $0.50$ & $0.79$ & $0.71$ & $0.91$\\
\hline
\end{tabular}
\end{center}
\caption{{\labelstyle Compares optimization methods when using min-margins constraints without Hedgehog shape prior. $X$ implies that the corresponding label was not assigned to any pixel in the final solution. The precision and recall were averaged over 15 examples. Our method clearly outperformed QPBO and \a-exp.
On average QPBO left $8.5\%$ of the pixels unlabeled and in one instance $11.5\%$.}}
\label{tbl:brainnohhog}
\vspace{-1ex}
\end{table*}

\noindent\textbf{Scenario II:} when $\alpha \not\in \Tree(A)$:

Case 1, assume $\clabelvar_p \in \Tree(A)$ and $\clabelvar_q \in \Tree(A)$. This case is similar to scenario I, case 1, the added edge is shown in Fig.~\ref{fig:pathmoveshapepriorconst}(b).

Cases 2, assume $\clabelvar_p \not \in \Tree(A)$. We can deduce that $\gpath[\clabelvar_p,\alpha] \not \subseteq \Tree(A)$. Thus, no edge needed since $\labelvar_p$ can not be in $\Tree(A)$.

Case 3, assume $\clabelvar_p \in \Tree(A)$ and $\clabelvar_q \not \in \Tree(A)$,  see case 3 above.

\vspace{-1ex}
\section{Experiments}
\vspace{-1ex}
\label{sc:experiments}
Our 2D medical segmentation experiments focus on comparing Path-Moves for optimizing energy \eqref{eqn:OurMainEnergy} or \eqref{eqn:OurMainEnergy}+\eqref{eq:shapprior} to QPBO \cite{rotherQPBO,DB:ICCV09} and \a-exp \cite{BVZ:PAMI01,DB:ICCV09}. In all experiments $\lambda$ was set to 1. To define our tree-metric, every edge $(\gamma,\beta)$ in $\Tree$ was assigned a non-negative weight $\pairwise_{pq}(\gamma,\beta)$ computed using a non-increasing function of difference in $p$ and $q$ intensities similar to \cite{BF:IJCV06}.
Also, whenever a Hedgehog \cite{isackhedgehog} shape prior was used its tightness parameter was set to $\pi/9$.

The experiments evaluate the effectiveness of Path-Moves as a combinatorial multi-labeling move. As such we assume that the color models are known a priori. One can easily integrate Path-Moves in a framework that estimates initial color models using user interaction and iteratively alternates between labeling pixels and re-estimating color models in a GrabCut fashion, e.g.~\cite{GrabCuts:SIGGRAPH04,LabelCosts:IJCV12,PEARL:IJCV12}.

\begin{figure}
\begin{center}
\begin{tabular}{@{\hspace{0ex}}c@{\hspace{0ex}}c@{\hspace{1ex}}c@{\hspace{1ex}}c@{\hspace{1ex}}c}
\multicolumn{5}{c}{\includegraphics[width=0.60\linewidth]{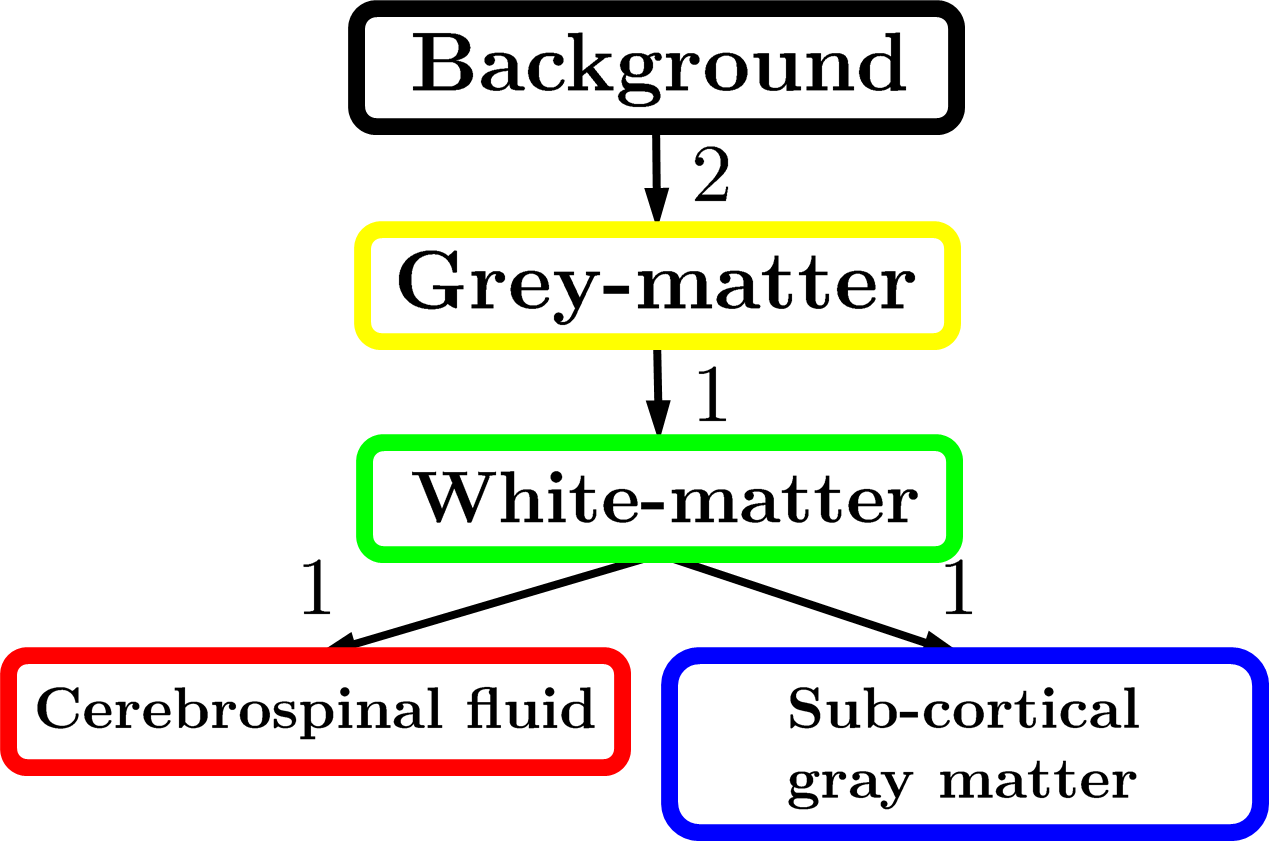}}\\
\multicolumn{5}{c}{ (a) tree and min-margins }\\
\begin{sideways}\phantom{xxx}Subject 2\end{sideways}&
\includegraphics[height=0.1\textheight,width=0.22\linewidth]{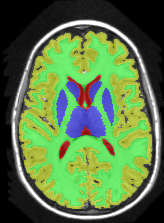}&
\includegraphics[height=0.1\textheight,width=0.22\linewidth]{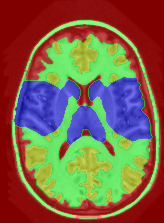}&
\includegraphics[height=0.1\textheight,width=0.22\linewidth]{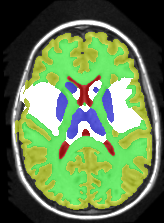}&
\includegraphics[height=0.1\textheight,width=0.22\linewidth]{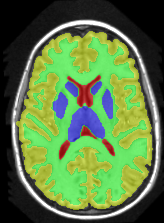}\\
\begin{sideways}\phantom{xxx}Subject 3\end{sideways}&
\includegraphics[height=0.1\textheight,width=0.22\linewidth]{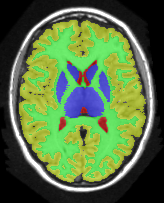}&
\includegraphics[height=0.1\textheight,width=0.22\linewidth]{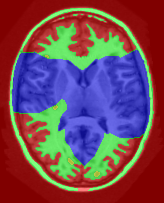}&
\includegraphics[height=0.1\textheight,width=0.22\linewidth]{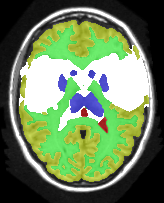}&
\includegraphics[height=0.1\textheight,width=0.22\linewidth]{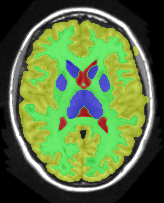}\\
\begin{sideways}\phantom{xxx}Subject 4\end{sideways}&
\includegraphics[height=0.1\textheight,width=0.22\linewidth]{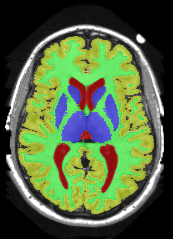}&
\includegraphics[height=0.1\textheight,width=0.22\linewidth]{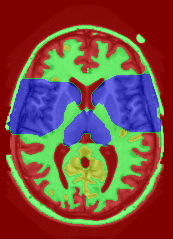}&
\includegraphics[height=0.1\textheight,width=0.22\linewidth]{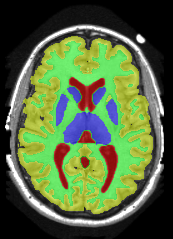}&
\includegraphics[height=0.1\textheight,width=0.22\linewidth]{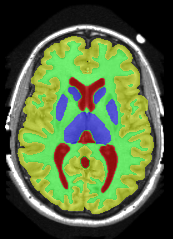}\\
&ground truth &\a-exp \cite{BVZ:PAMI01,DB:ICCV09} & QPBO \cite{rotherQPBO,DB:ICCV09} & ours
\end{tabular}
\end{center}
\caption{{\labelstyle sample results when using tree in (a). An arc weight in (a) represents the min-margin of the head node. White pixels were unlabeled by QPBO. Path-Moves outperformed QPBO is most cases, and \a-exp in all cases.}}
\label{fig:brainresults}
\vspace{-2.5ex}
\end{figure}

\vspace{-3ex}
\paragraph{Brain Segmentation:} We combined the labeled regions in dataset \cite{braindataset} (T1W MRI) to create the tree shown in Fig.~\ref{fig:brainresults}(a). In this setting, the data term is the sum of color model penalty and an $L_2$ shape prior \cite{pdecuts:ECCV06} based on an automatically extracted brain mask using \cite{jenkinson2005BrainMask},
\begin{equation}
\!\!\!\!\unary_p(\labelvar_p)=\!\left\{
\begin{array}{ll}
\!\!\!- \ln(Pr(I_p|\labelvar_p)) & \text{background}\\
\!\!\!- \ln(Pr(I_p|\labelvar_p))+DT(p) & \text{otherwise,}
\end{array}
\right.
\end{equation}
where $I_p$ is the intensity at pixel $p$ and $DT$ is the Euclidean Distance Transform of the extracted brain mask. Min-margins are shown in Fig.~\ref{fig:brainresults}(a). We also added a Hedgehog prior \cite{isackhedgehog} for the sub-cortical grey-matter to help our energy differentiate between grey-matter and sub-cortical grey-matter.

In this application our method outperformed QPBO in most cases and \a-exp in all cases. In fact \a-exp  always converged to a weak local minima in this setting, see Fig.~\ref{fig:brainresults}. Based on our experience the quality of \a-exp result depends on various factors, e.g.~tree complexity, the number of min-margins introduced, the order in which labels are expanded, and the initial solution. For the subjects that QPBO was able to find the global optimal Path-Moves either found the global optimal or a very close solution.

\newcommand{\priorfigh}{0.15}
\begin{figure}[h]
\begin{center}
\begin{tabular}{@{\hspace{0ex}}c@{\hspace{0ex}}c@{\hspace{0ex}}c@{\hspace{1ex}}c@{\hspace{1ex}}c@{\hspace{1ex}}c}
\begin{sideways} \phantom{xxx} min-margins\end{sideways}&
\begin{sideways} \phantom{xx} Hedgehog prior\end{sideways}&
\begin{sideways} \phantom{xxx}\end{sideways}&
\includegraphics[height=\priorfigh\textheight]{m_1210000_alphaexp.png}&
\includegraphics[height=\priorfigh\textheight]{m_1210000_qpbo.png}&
\includegraphics[height=\priorfigh\textheight]{m_1210000_pathmoves.png}\\
\begin{sideways} \phantom{xxx} min-margins\end{sideways}&
\begin{sideways} \phantom{xx}\textbf{No} Hedgehog prior\end{sideways}&
\begin{sideways} \phantom{xxx}\end{sideways}&
\includegraphics[height=\priorfigh\textheight]{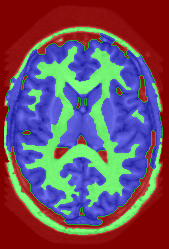}&
\includegraphics[height=\priorfigh\textheight]{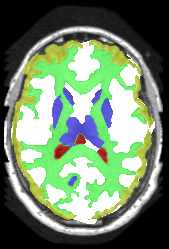}&
\includegraphics[height=\priorfigh\textheight]{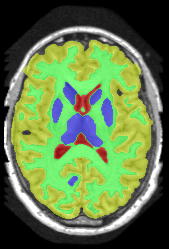}\\
\begin{sideways} \phantom{xx}\textbf{No} min-margins\end{sideways}&
\begin{sideways} \phantom{xx}Hedgehog prior\end{sideways}&
\begin{sideways} \phantom{xxx}\end{sideways}&
\includegraphics[height=\priorfigh\textheight]{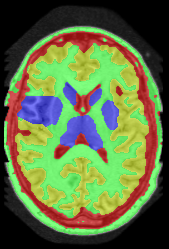}&
\includegraphics[height=\priorfigh\textheight]{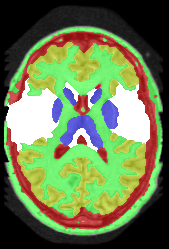}&
\includegraphics[height=\priorfigh\textheight]{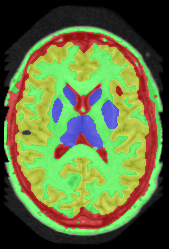}\\
\begin{sideways} \phantom{xxx}\textbf{No} min-margin\end{sideways}&
\begin{sideways} \phantom{x}\textbf{No} Hedgehog prior\end{sideways}&
\begin{sideways} \phantom{xxx}\end{sideways}&
\includegraphics[height=\priorfigh\textheight]{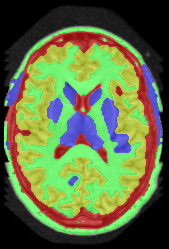}&
\includegraphics[height=\priorfigh\textheight]{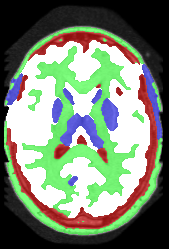}&
\includegraphics[height=\priorfigh\textheight]{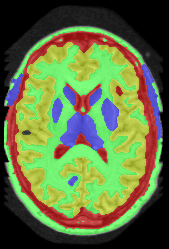}\\
&&&\a-exp \cite{BVZ:PAMI01,DB:ICCV09} & QPBO \cite{rotherQPBO,DB:ICCV09} & ours\\[-1.5ex]
\end{tabular}
\end{center}
\caption{{\labelstyle show Subject 1 results with/without min-margins and/or Hedgehog shape prior. The first column indicates whether min-margins and/or Hedgehog where used or not.}}
\label{fig:brainresultszm}
\vspace{-2ex}
\end{figure}

Figure \ref{fig:brainresultszm} shows the results for Subject 1 with (and without) min-margins and Hedgehog prior. The third row shows the results when not using min-margins. Path-Moves converged after two iterations to a lower energy than \a-exp, which converged after six iterations. In this case \a-exp local minimum was due to the Hedgehog prior, see last row.

Table \ref{tbl:brainhhog} compares the precision, recall and F$_1$ score for each region individually, where
$F_1=2\frac{precision\cdot recall}{precision\;+\;recall}$. The higher F$_1$ values correspond to better segmentation. \linebreak
In general, QPBO was unpredictable as in some cases it found the optimal solution and in other cases it left a large number of pixels unlabeled.

Table \ref{tbl:brainnohhog} show the results after dropping the hedgehog prior. In terms of Path-Moves, the mainly affected label after dropping the Hedgehog prior is the sub-cortical gray matter, as it started to grab parts of the gray matter, see Fig.\ref{fig:brainresultszm} second row, last column. Comparing Tables \ref{tbl:brainhhog} and \ref{tbl:brainnohhog} it is clear that QPBO and \a-exp benefited the most by introducing the hedgehog prior.

\begin{figure}[h]
\begin{center}
\begin{tabular}{c@{\hspace{1.5ex}}c}
\multicolumn{2}{c}{\includegraphics[width=0.5\linewidth]{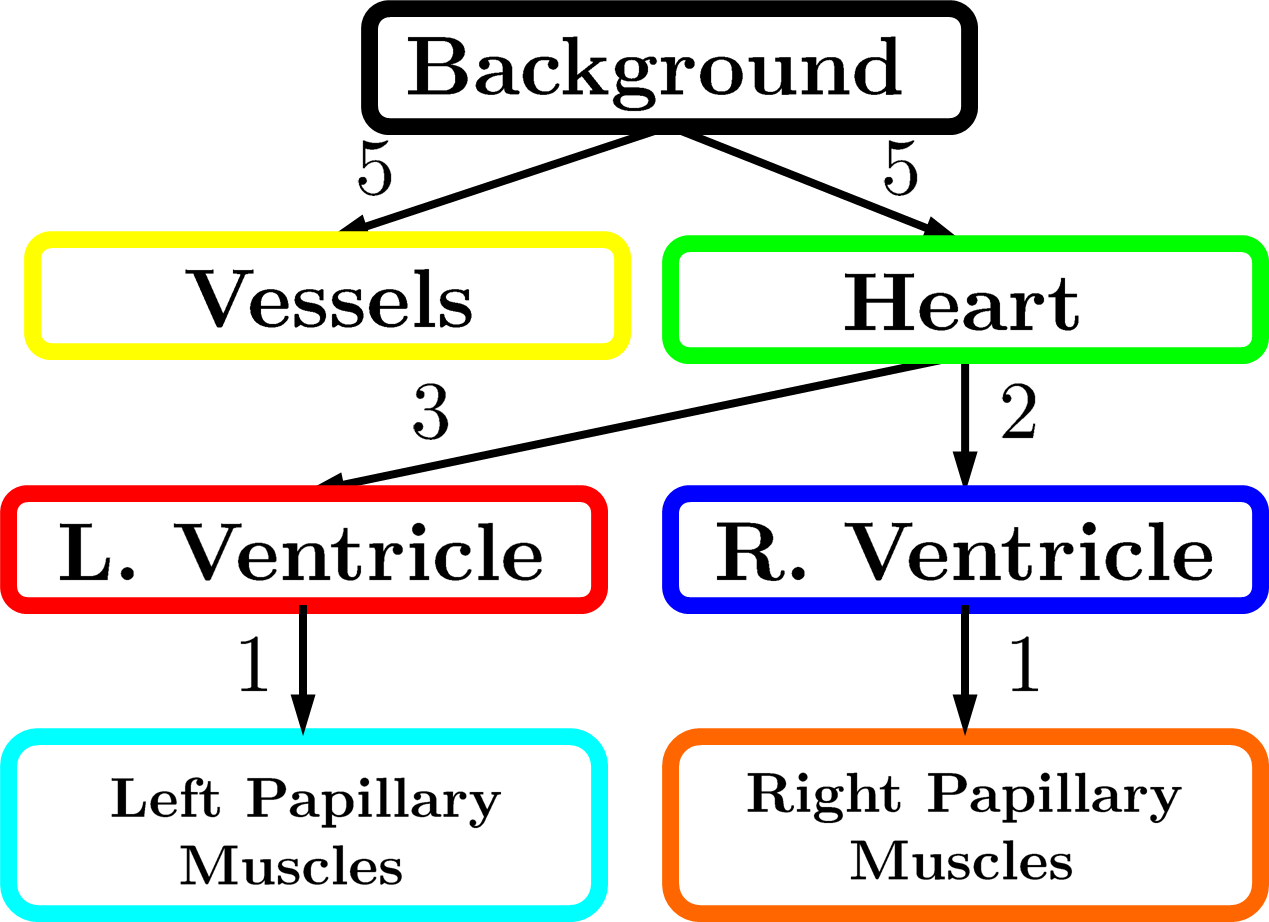}}\\
\multicolumn{2}{c}{ (a) heart tree}\\
\includegraphics[height=0.13\textheight,width=0.43\linewidth]{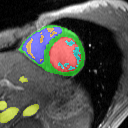}&
\includegraphics[height=0.13\textheight,width=0.43\linewidth]{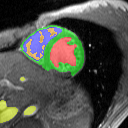}\\
{\small (b) ground truth}&{\small (c) \a-exp\cite{BVZ:PAMI01,DB:ICCV09}}\\
\includegraphics[height=0.13\textheight,width=0.43\linewidth]{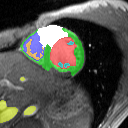}&
\includegraphics[height=0.13\textheight,width=0.43\linewidth]{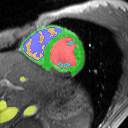}\\
{\small (d) QPBO \cite{rotherQPBO,DB:ICCV09}} & {\small (e) ours (Path-Moves)}\\[-1.5ex]
\end{tabular}
\end{center}
\caption{{\labelstyle Heart segmentation using tree shown in (a). Using \a-exp leads to a local minimum. Path-Moves avoids this minimum via multi-label expansions. QPBO left many pixels unlabeled.}}
\label{fig:heartresult}
\end{figure}

\vspace{-4ex}
\paragraph{Heart Segmentation:} In this setting we only used color models for the data term and no shape priors. Figure \ref{fig:heartresult}(a) shows the labels tree. For \a-exp to escape its local minimum it needs to first expand the left ventricle and then the left papillary muscles. However, expanding on left ventricle would lead to a higher energy than the current one.
Path-Moves avoids this local minimum by allowing both labels to expand simultaneously when performing a Path-Move on the left papillary muscles.

\vspace{-2ex}
\paragraph{Abdominal Organ Segmentation:} We used CT dataset and extended the work in \cite{isackhedgehog} which used Hedgehogs to segment liver and kidneys. In contrast to \cite{isackhedgehog}, we utilized more detailed structures reaching $13$ labels.

\begin{table}[h]
\begin{center}
\begin{tabular}{l|c|c|c|}
\cline{2-4}
& Ours & QPBO  & \a-exp \\
\hline
\multicolumn{1}{|c|}{F$_1$ Score}& $\mathbf{0.95}$ & $\mathbf{ 0.95}$ & $0.93$\\
\hline
\multicolumn{1}{|c|}{Weighted Precision}& $0.95$ & $0.95$ & $0.94$ \\
\hline
\multicolumn{1}{|c|}{Weighted Recall} & $0.95$ & $0.95$ & $0.92$\\
\hline
\end{tabular}
\end{center}
\vspace{-1ex}
\caption{{\labelstyle Weighted prec. and recall were averaged over 7 test cases. All methods preformed comparably when using Hedgehogs \cite{isackhedgehog}.
}}
\label{tbl:organshhog}
\end{table}

\begin{table}[h]
\begin{center}
\begin{tabular}{l|c|c|c|}
\cline{2-4}
& Ours & QPBO  & \a-exp \\
\hline
\multicolumn{1}{|c|}{F$_1$ Score}& $\mathbf{0.88}$ & $0.85$ & $0.82$\\
\hline
\multicolumn{1}{|c|}{Weighted Precision}& $0.92$ & $0.93$ & $0.89$ \\
\hline
\multicolumn{1}{|c|}{Weighted Recall} & $0.85$ & $0.78$ & $0.76$\\
\hline
\end{tabular}
\end{center}
\caption{{\labelstyle Weighted precision and recall were averaged over 7 test cases. Eliminating Hedgehog priors \cite{isackhedgehog} lead to a significant decrease in segmentation quality.
On average QPBO left $10.8\%$ of the pixels unlabeled.}
}
\label{tbl:organsnohhog}
\end{table}

\begin{figure}[h]
\begin{center}
\begin{tabular}{cc}
\multicolumn{2}{c}{ \includegraphics[width=0.9\linewidth]{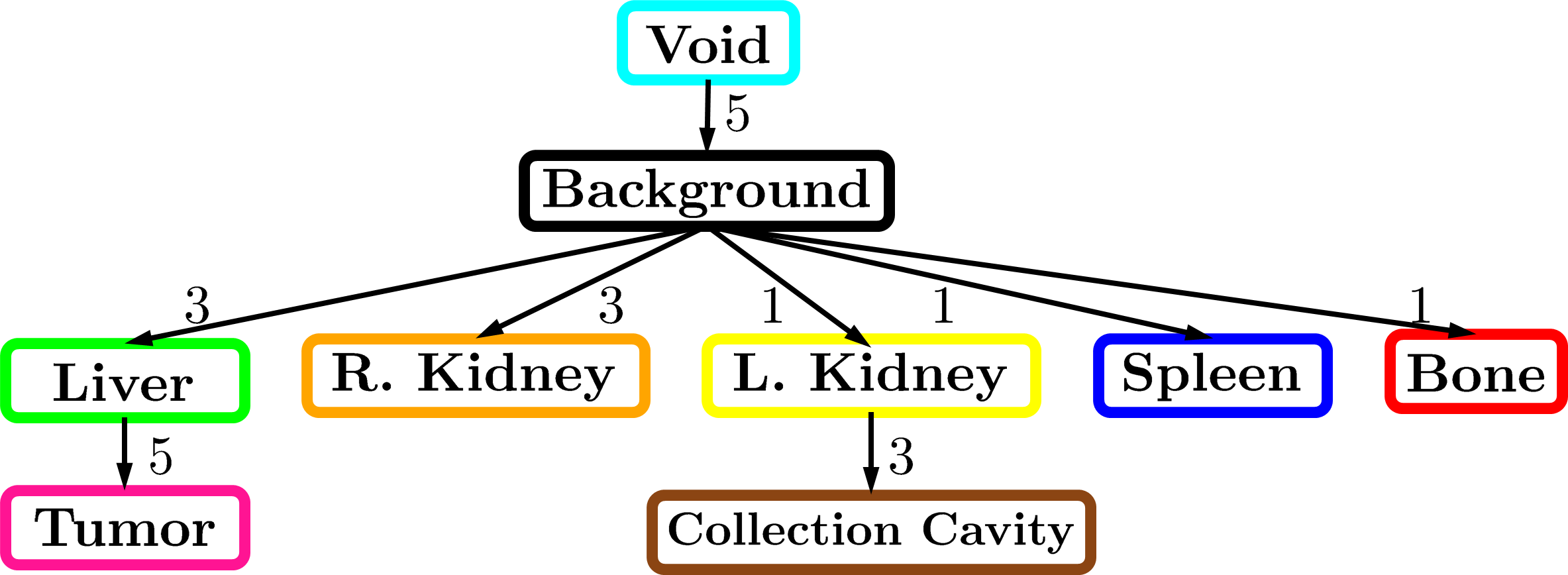}}\\
\multicolumn{2}{c}{ (a) tree}\\
\includegraphics[width=0.45\linewidth]{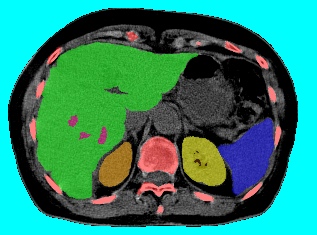}&
\includegraphics[width=0.45\linewidth]{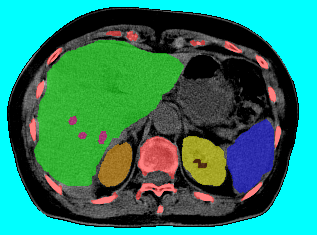}\\
(b) ground truth & (c) ours (Path-Moves)
\end{tabular}
\end{center}
\caption{{\labelstyle(a) abdominal organs structure used for this example. Void is the empty space around the body. We only show our result as QPBO and \a-exp results were almost identical to ours.}}
\label{fig:organsresult}
\end{figure}

For each test case we computed the weighted precision
\vspace{-1ex}
$$\text{weighted precision} = \sum_{\ell \in \labelset} \frac{|\labelvars^*=\ell|}{|\pixelset|} \text{precision}_\ell
\vspace{-1ex}
$$ where $\labelvars^*$ is the ground truth labeling. The weighted recall is defined similarly. As shown in Table \ref{tbl:organshhog}, all methods performed comparably due to the use of Hedgehog priors and the star-like structure of $\Tree$, which $\a$-exp is well suited for. See Table \ref{tbl:organsnohhog} for results without using Hedgehog priors.
Figure \ref{fig:organsresult} shows the tree and our result for one test case. Interestingly, QPBO labeled all the pixels in all 7 test cases. By comparing Tables \ref{tbl:organshhog} and \ref{tbl:organsnohhog} it is easy to see the benefit of using Hedgehog priors. Moreover, Path-Moves outperformed QPBO and \a-exp after dropping the Hedgehog priors.

\begin{figure}[h]
\begin{center}
\begin{tabular}{@{\hspace{0ex}}c@{\hspace{0.5ex}}c}
\multicolumn{2}{c}{ \includegraphics[width=0.75\linewidth]{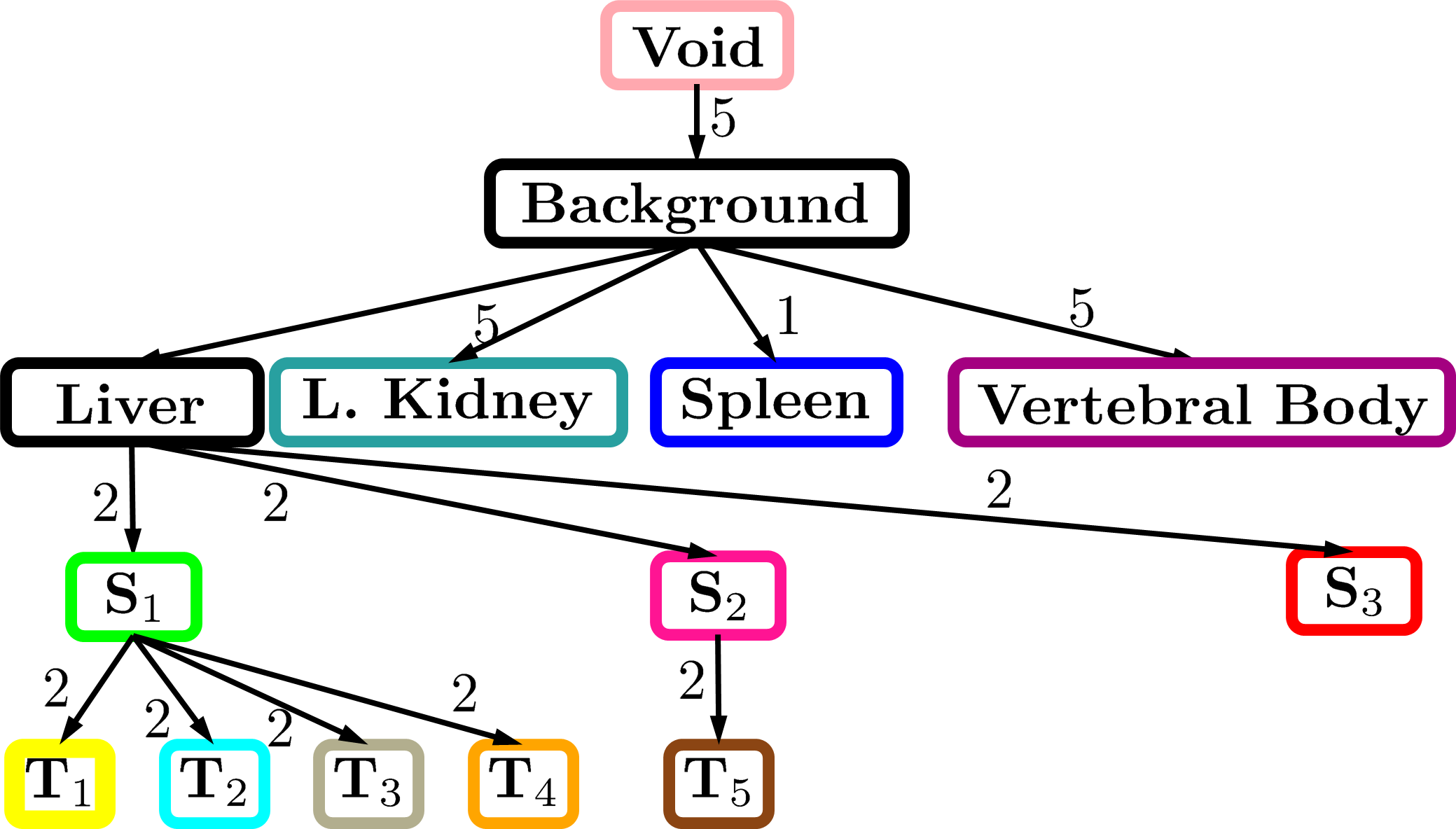}}\\
\multicolumn{2}{c}{ (a) tree}\\
\includegraphics[width=0.5\linewidth]{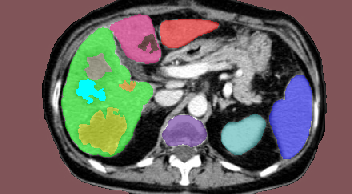}&
\includegraphics[width=0.5\linewidth]{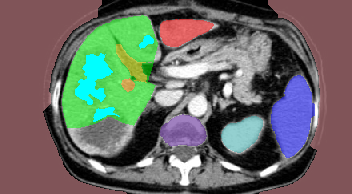}\\
(b) ground truth & (c) \a-exp\\
\includegraphics[width=0.5\linewidth]{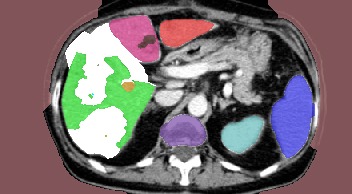}&
\includegraphics[width=0.5\linewidth]{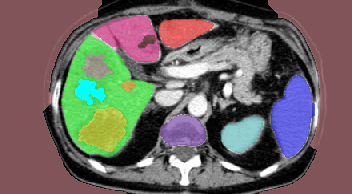}\\
(d) QPBO & (e) ours (Path-Moves)
\end{tabular}
\end{center}
\caption{{\labelstyle(a) a challenging abdominal organs structure. $S_x$ and $T_y$ denote liver segment x and tumor y, respectively. The liver label in (a) is a conceptual/artificial label with infinity data term penalty. Our method significantly outperformed QPBO and \a-exp.
}}
\label{fig:lobedliverresult}
\end{figure}

We pursued a more challenging structure, see Fig.\ref{fig:lobedliverresult}(a). The objective in this case was to segment the liver into three segments and any tumors inside them separately. Due to the large overlap between color models and the complex structure, having Hedgehog priors was not enough for QPBO or \a-exp to converge to an adequate solution, see Fig.\ref{fig:lobedliverresult}(c-e).
Path-Moves was able to achieve good results by avoiding local minima as in Fig.\ref{fig:lobedliverresult}(c). Furthermore, Path-Moves guarantees full labeling in contrast to QPBO, which left $7.4\%$ of the pixels unlabeled in Fig.\ref{fig:lobedliverresult}(d).

In conclusion, our results empirically show that for a general tree \a-exp converges to weak local minima, and QPBO is unpredictable in terms of being able to label all pixels. Path-Moves which uses an effective multi-label expansion move, labels all pixels and easily avoids weak local minima that \a-exp is prone to.

\section{Discussion}
\label{sc:discussion}
Path-Moves is applicable to tree-metrics which could be used to approximate arbitrary metrics \cite{Kumar:2009,PedroGobalTreeMetric}. Even in the absence of interactions, Path-Moves is a more powerful move making algorithm than \a-exp \cite{BVZ:PAMI01} because of the {\em multi-label} nature of its moves. Thus, Path-Moves is a better fit for applications that rely on tree-metrics such as \cite{Kumar:2009}.
In the presence of interaction constraints the optimality bound of \cite{BVZ:PAMI01} is not valid. The proof in \cite{BVZ:PAMI01} assumes that given any labeling every pixel with ground truth label $X$ could switch to $X$ via a binary expansion on $X$. This is no longer guaranteed as interaction constraints limit \cite{BVZ:PAMI01} expansion domain, e.g.~see Fig.\ref{fig:aexplocalminima}(c). Our experiments empirically show that Path-Moves finds optimal or near optimal solution.
In the cases where QPBO found full labeling, i.e.~optimal solution, Path-Moves either found the same solution or a very close one, see Table \ref{tbl:organshhog} and Fig.\ref{fig:brainresults} Subject 4.

In terms of space complexity \a-exp is the most efficient as it requires building a graph with $O(|\pixelset|)$ nodes while QPBO requires a significantly larger graph with $O(|\pixelset||\labelset|)$ nodes. A Path-Move graph size depends on $\Tree$. When $\Tree$ is balanced it requires $O(|\pixelset|\ln(|\labelset|))$ nodes and $O(|\pixelset||\labelset|)$ in the worse case when $\Tree$ is a chain.

There is one limitation when using our Path-Moves to optimize \eqref{eqn:OurMainEnergy} compared to \cite{DB:ICCV09}. In \cite{DB:ICCV09} it is possible to explicitly control the min.~exclusion margin between two siblings, say A and B in $\Tree$. In our model the min.~exclusion margin is implicit and it is equal to $\max(\margins_A,\margins_B).$ Because siblings such as $A$ and $B$ are not directly connected in the tree.

Another limitation are interaction constraints that are not Path-Move representable. An interaction constraint is not Path-Move representable if there exists $\alpha, \beta$ and $\gamma \in {\cal L}$
where $a< b \in \Gamma(\gamma,\alpha)$ and $c< d \in \Gamma(\beta,\alpha)$ while
configuration $[a,d]$ is prohibited and $[b,c]$ is permissible \cite{schlesinger2006transforming}.
In general, this could be avoided either by slightly modifying tree $\cal T$ or relaxing the interaction constraints.

\begin{figure*}[ht]
\begin{center}
\begin{tabular}{@{\hspace{0ex}}c@{\hspace{0.5ex}}c@{\hspace{0ex}}c@{\hspace{0ex}}c}
\includegraphics[align=c,width=0.26\linewidth]{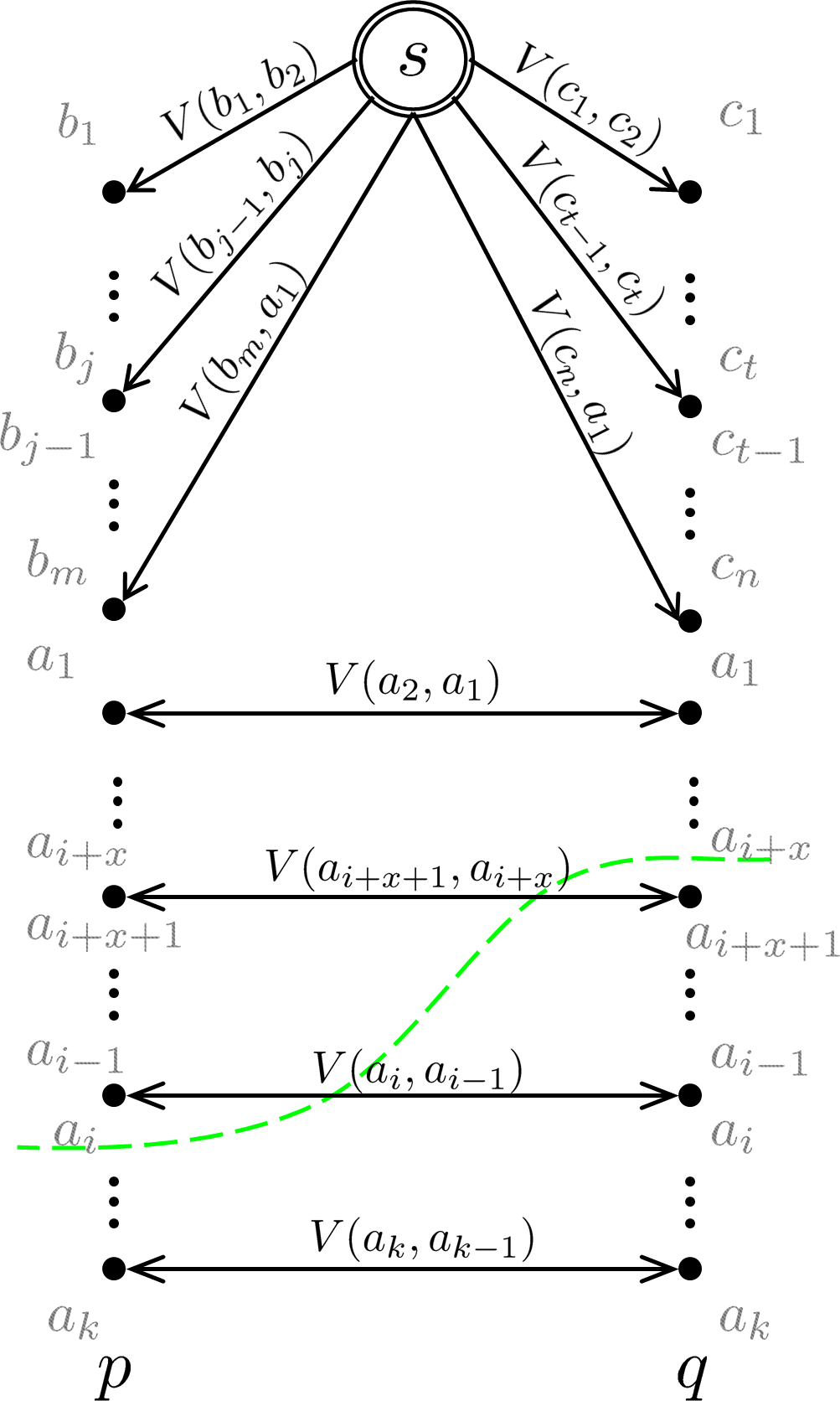}&
\includegraphics[align=c,width=0.26\linewidth]{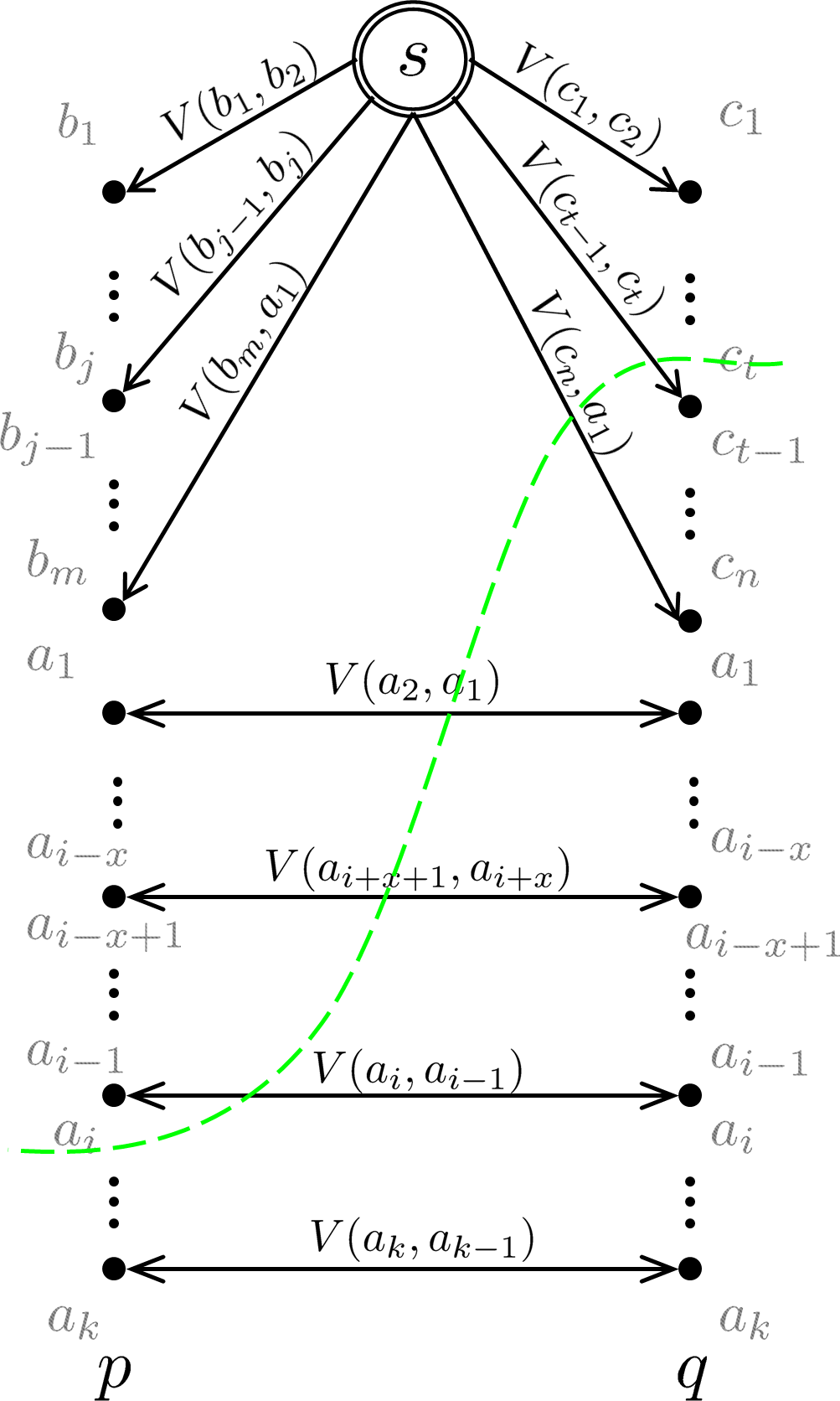}&
\includegraphics[align=c,width=0.26\linewidth]{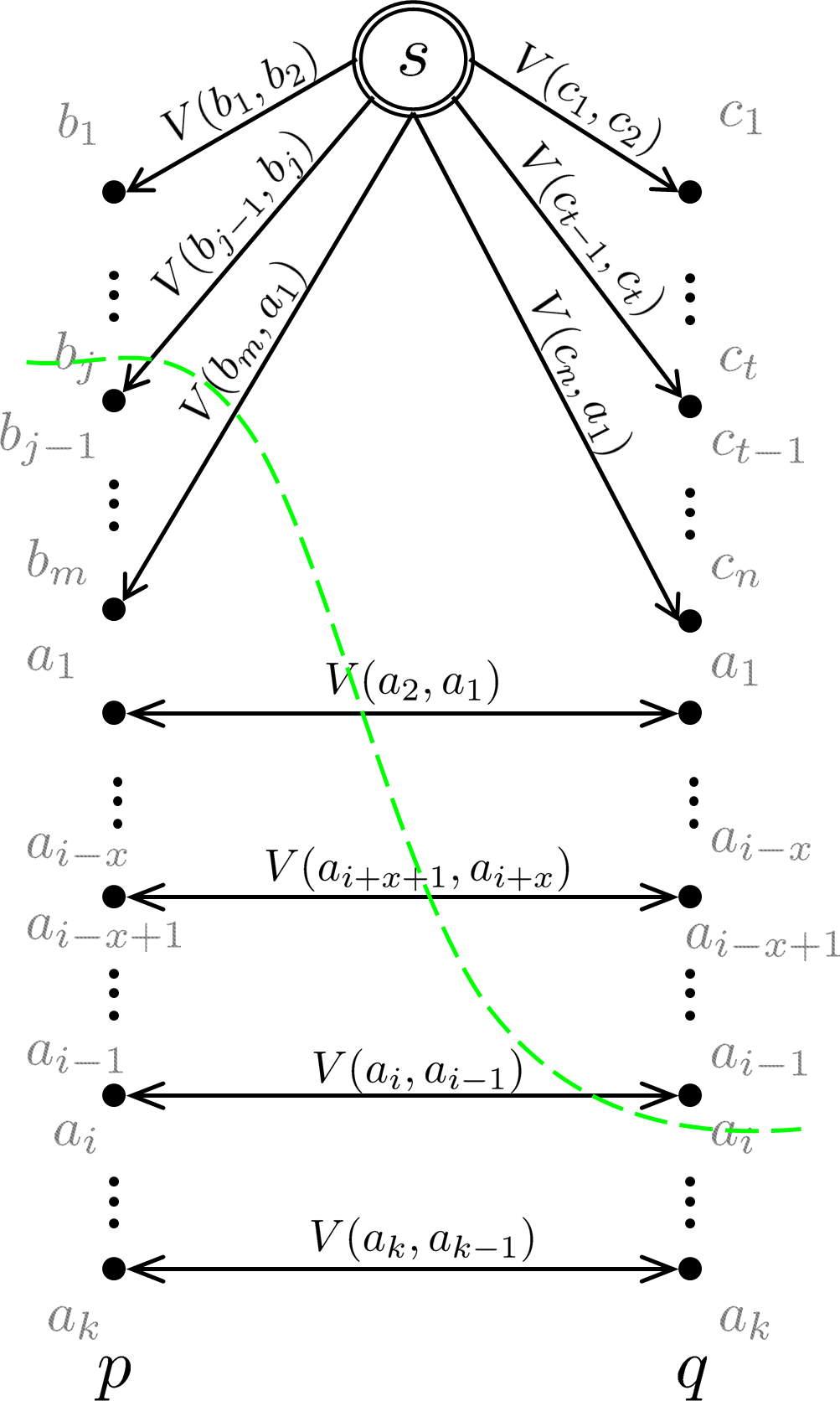}&
\includegraphics[align=c,width=0.26\linewidth]{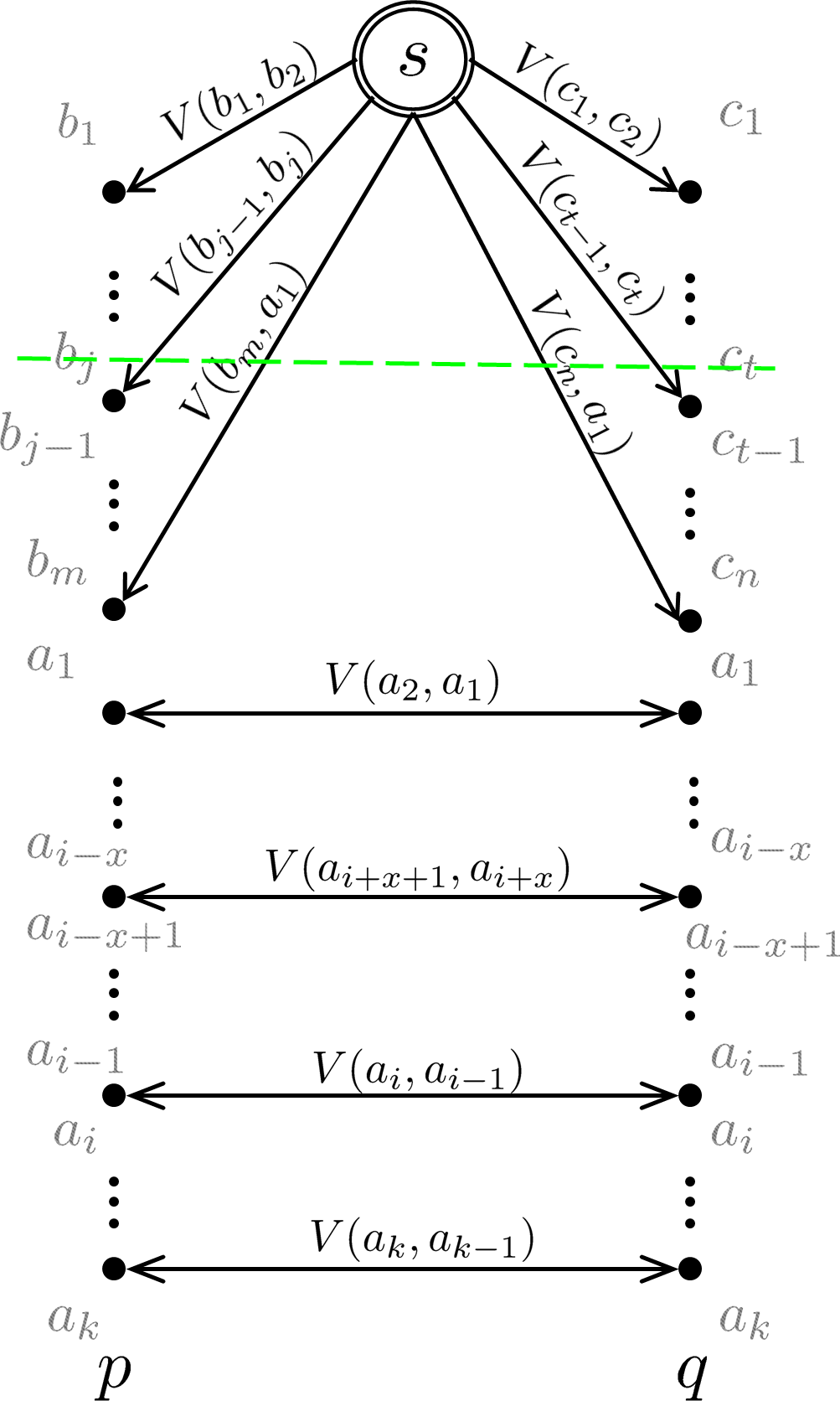}\\
{\small (a) $f_p=a_i,$ $f_q=a_{i+x}$} &
{\small (b) $f_p=a_i,$ $f_q=c_t$} &
{\small (c) $f_p=b_j,$ $f_q=a_i$} &
{\small (d) $f_p=b_j,$ $f_q=c_t$}
\end{tabular}
\end{center}
\caption{{\labelstyle shows four general cuts representative of all possible cuts involving pixels $p$ and $q$. Note that along each expansion path we could only cut one edge due to the backward $w_\infty$ edges, which were introduced for encoding data terms to enforce exclusion.}}
\label{fig:vcorrectness}
\end{figure*}

\vspace{-1ex}
\section{Conclusion}
\vspace{-1ex}
\label{sc:conclusion}
The proposed multi-labeling move is effective in optimizing models with  hierarchically-structured segments (partially ordered labels) and interaction constraints. In contrast to binary expansion move \cite{BVZ:PAMI01}, our move avoids local minima caused by interaction constraints.

Our experiments cover various medical segmentation applications, e.g.~brain and heart segmentation.
Our results show that Path-Moves always perform at least as well as prior methods. Moreover, Path-Moves significantly outperform prior methods when using complex trees and/or regions with ambiguous color models.

Path-Moves is applicable to arbitrary trees. This is in contrast to \cite{DB:ICCV09} which is not easy to generalize for an arbitrary tree as it relies on the cumbersome process of reducing high-order data terms to unary and pairwise potentials.


We generalized star-like shape priors in the context of partially ordered labels. Extending preexisting commonly used priors to partially ordered labels is an interesting idea on its own and we leave this for future work.

\vspace{-1ex}
\paragraph{Acknowledgments}
\label{sc:acknowledgments}
This work was supported by NIH grants R01-EB004640, P50-CA174521 and R01-CA167632. We thank Drs.~S. O'Dorisio and Y. Menda for providing the liver data (NIH grant U01-CA140206).
This work was also supported by NSERC Discovery and RTI grants (Canada) for Y. Boykov and O. Veksler.

\appendix
\section{Smoothness Encoding Proof Of Correctness} \label{ap:smoothnessproof}
The following proof uses some of the tree-metric function $V$ properties, mainly symmetry and tree-metric condition \eqref{eqn:treemetriccond}.
Let $p$ and $q$ be a pair of neighbour pixels and assume that we are expanding on label $\alpha \in \labelset$. Recall that our graph construction treats the sequence of overlapping labels of paths $\gpath[\clabelvar_p,\alpha]$ and $\gpath[\clabelvar_q,\alpha]$ differently from the non-overlapping parts.
Therefore, we rename $\gpath[\clabelvar_p,\alpha]=(b_1,\hdots,b_m,a_1,\hdots,a_k)$ and $\gpath[\clabelvar_q,\alpha] =(c_1,\hdots,c_n,a_1,\hdots,a_k)$ to emphasize the overlap.

To show the correctness of our smoothness encoding shown in Fig.\ref{fig:pathmoveconstruction}(b) we need to consider the cost of all possible cuts involving pixels $p$ and $q$. However, it is enough to only consider the four general cuts shown in Fig.~\ref{fig:vcorrectness}, which are representative of all possible cuts.

\textbf{Case I}, $\labelvar_p \in (a_1,\hdots,a_k)$ and $\labelvar_q \in (a_1,\hdots,a_k)$: Let $\labelvar_p=a_i$ and $\labelvar_q=a_{i+x}$ where $1-i \leq x \leq k-i$.
If $1-i \leq x < 0$ then the cost of severed edges, shown in Fig.~\ref{fig:vcorrectness}(a), will be
\begin{multline}
\label{eq:costaa}
cost=V(a_{i},a_{i-1})+ V(a_{i-1},a_{i-2}) + \hdots\\ +V(a_{i+x+1},a_{i+x}).
\end{multline}
Using equation \eqref{eqn:treemetriccond} we can deduce that the terms in \eqref{eq:costaa} sum to $V(a_i,a_{i+x})$, which is the correct smoothness cost based on our assumption regarding $\labelvar_p$ and $\labelvar_q$.

If $x=0$ then based on our construction no smoothness edges will be severed between that the $p$ and $q$ expansion paths. Thus, according to our construction the cost for the assigning $p$ and $q$ to $a_i$ will be 0, which is correct since $\pairwise(\ell,\ell)=0$ for all $\ell \in \labelset$ by definition.

Finally, the proof when  $0<x \leq k-i$ is equivalent to changing the roles of $p$ and $q$ when $1-i \leq x <0 $.

\textbf{Case II}, $\labelvar_p \in (a_1,\hdots,a_k)$ and $\labelvar_q \in (c_1,\hdots,c_n)$: Let $\labelvar_p=a_i$ and $\labelvar_q=c_t$.
The cost of severed edges, shown in Fig.\ref{fig:vcorrectness}(b), will be
\begin{multline*}
cost=V(a_{i},a_{i-1})+ \hdots +V(a_{2},a_{1})\\
    +V(a_1,c_n)+ \hdots +V(c_{t-1},c_{t}).
\end{multline*}
Using equation \eqref{eqn:treemetriccond} we can deduce the following
\begin{equation*}
\begin{split}
cost= &\overbrace{V(a_{i},a_{i-1}) +V(a_{2},a_{1})}^{V(a_i,a_1)}+\hdots\\
    &\quad \quad + \underbrace{V(a_1,c_n)+ \hdots +V(c_{t-1},c_{t})}_{V(a_1,c_t)}\\
    =&V(a_i,c_t),
\end{split}
\vspace{-2ex}
\end{equation*}
which is the correct smoothness cost based  on our assumption regarding $\labelvar_p$ and $\labelvar_q$.

\textbf{Case III}, $\labelvar_p \in (b_1,\hdots,b_m)$ and $\labelvar_q \in (a_1,\hdots,a_k)$: Let $\labelvar_p=b_j$ and $\labelvar_q=a_i$. Using equation \eqref{eqn:treemetriccond} we can deduce that the cost of severed edges, shown in Fig.\ref{fig:vcorrectness}(c), will be
\begin{equation*}
\begin{split}
cost=&\overbrace{V(b_{j},b_{j-1})+ \hdots +V(b_{m},a_{1})}^{V(b_j,a_1)}\\
    &\quad\quad+ \underbrace{V(a_1,a_2)+ \hdots +V(a_{i-1},a_{i})}_{V(a_1,a_i)}\\
    =&V(b_j,a_i),
\end{split}
\end{equation*}
which is the correct smoothness cost based  on our assumption regarding $\labelvar_p$ and $\labelvar_q$.

\textbf{Case IV}, $\labelvar_p \in (b_1,\hdots,b_m)$ and $\labelvar_q \in (c_1,\hdots,c_n)$: Let $\labelvar_p=b_j$ and $\labelvar_q=c_t$. Using equation \eqref{eqn:treemetriccond} we can deduce that the cost of severed edges, shown in Fig.\ref{fig:vcorrectness}(d), will be
\begin{equation*}
\begin{split}
cost=&\overbrace{V(b_{j},b_{j-1})+ \hdots +V(b_{m},a_{1})}^{V(b_j,a_1)}\\
    &\quad\quad+ \underbrace{V(a_1,c_n)+ \hdots +V(c_{t-1},c_{t})}_{V(a_1,c_t)}\\
    =&V(b_j,c_t),
\end{split}
\end{equation*}
which is the correct smoothness cost based  on our assumption regarding $\labelvar_p$ and $\labelvar_q$.

Recall that for any pixel $p$ a non-prohibitively expensive min.~cut severs only one edge along its corresponding graph chain $(s,{\mathcal C}_p,t)$, which is due to the data terms exclusion constraints.
Thus, this proof showed that our smoothness cost encoding is correct for any feasible cut involving a pair of neighboring pixels.

{\small
\bibliographystyle{ieee}
\bibliography{pathmoves}
}

\end{document}